\documentclass[final,onefignum,onetabnum]{siamonline250211}

\usepackage{lipsum}
\usepackage{amsfonts}
\usepackage{graphicx}
\usepackage{epstopdf}
\usepackage{algorithmic}
\ifpdf
  \DeclareGraphicsExtensions{.eps,.pdf,.png,.jpg}
\else
  \DeclareGraphicsExtensions{.eps}
\fi

\usepackage{enumitem}
\setlist[enumerate]{leftmargin=.5in}
\setlist[itemize]{leftmargin=.5in}

\newsiamremark{remark}{Remark}
\newsiamremark{hypothesis}{Hypothesis}
\crefname{hypothesis}{Hypothesis}{Hypotheses}
\newsiamthm{claim}{Claim}
\newsiamremark{fact}{Fact}
\crefname{fact}{Fact}{Facts}

\usepackage{url}

\usepackage{graphicx,xcolor}
\usepackage{booktabs}
\usepackage{subcaption}
\usepackage{wrapfig}
\usepackage{pifont} 

\usepackage{amsmath}
\usepackage{amssymb}
\usepackage{mathtools}
\mathtoolsset{showonlyrefs}
\usepackage{bm}
\usepackage{tikz}
\usetikzlibrary{calc}

\newcommand{\cmark}{\textcolor{green!60!black}{\ding{51}}}%
\newcommand{\xmark}{\textcolor{red}{\ding{55}}}%

\def\1{\bm{1}}

\def\d{\mathrm{d}}

\def\bX{{\bm{X}}}

\def\fA{{\bm{A}}}

\def\bY{{\bm{Y}}}

\def\bZ{{\bm{Z}}}

\def\bH{{\bm{H}}}

\def\bW{{\bm{W}}}

\def\rv{{\textnormal{v}}}

\def\dd{{\mathrm{d}}}

\def\vc{{\bm{c}}}

\def\vx{{\bm{x}}}
\def\vy{{\bm{y}}}

\DeclareMathAlphabet{\mathsfit}{\encodingdefault}{\sfdefault}{m}{sl}
\SetMathAlphabet{\mathsfit}{bold}{\encodingdefault}{\sfdefault}{bx}{n}

\def\sP{{\mathbb{P}}}
\def\sQ{{\mathbb{Q}}}

\newcommand{\E}{\mathbb{E}}

\newcommand{\R}{\mathbb{R}}

\let\log\relax
\DeclareMathOperator{\log}{ln}

\def\P{{\mathbb{P}}}
\def\F{{\mathbb{F}}}

\def\R{{\mathbb{R}}}

\newcommand{\KL}{\mathrm{KL}}

\DeclareMathOperator*{\argmin}{arg\,min}

\definecolor{pearDark}{HTML}{2980B9}

\renewcommand{\c}[1]{\ensuremath{\mathbf{#1}}}    

\renewcommand{\rv}[1]{\ensuremath{\MakeUppercase{\c{#1}}}}  

\newcommand{\fwd}[1]{%
  \tikz[baseline=(char.base)]{
    \node[inner sep=0pt, outer sep=0pt] (char) {$#1$};
    \draw[line width=0.2pt] ($(char.north west)+(0.05em,0.25em)$) -- ($(char.north east)+(0em,0.25em)$);
    \draw[line width=0.2pt] ($(char.north east)+(0em,0.25em)$) -- ($(char.north east)+(-0.15em,0.15em)$);
  }%
} 

\newcommand{\bwd}[1]{%
  \tikz[baseline=(char.base)]{
    \node[inner sep=0pt, outer sep=0pt] (char) {$#1$};
    \draw[line width=0.2pt] ($(char.north west)+(0em,0.25em)$) -- ($(char.north east)+(-0.05em,0.25em)$);
    \draw[line width=0.2pt] ($(char.north west)+(0em,0.25em)$) -- ($(char.north west)+(0.15em,0.15em)$);
  }%
}

\headers{Iterative Importance Fine-tuning of Diffusion Models}{A. Denker, S. Padhy, F. Vargas, J. Hertrich}

\title{Iterative Importance Fine-tuning of Diffusion Models\thanks{AD acknowledges support from the EPSRC (EP/V026259/1). JH acknowledges funding by the Deutsche Forschungsgemeinschaft (DFG, German Research Foundation) within project no 530824055.}}

\author{Alexander Denker\thanks{University College London 
  (\email{a.denker@ucl.ac.uk}).}
\and
Shreyas Padhy\thanks{University of Cambridge 
  (\email{sp2058@cam.ac.uk}).}
  \and
Francisco Vargas\thanks{University of Cambridge \& Xaira Therapeutics  
  (\email{fav25@cam.ac.uk}).}
  \and 
  Johannes Hertrich\thanks{ENS Paris - PSL \& Inria Mokaplan
  (\email{johannes.hertrich@ens.fr}).}}

\usepackage{amsopn}

\ifpdf
\hypersetup{
  pdftitle={Iterative Importance Fine-tuning of Diffusion Models},
  pdfauthor={A. Denker, S.  Padhy, F. Vargas, J. Hertrich}
}
\fi

\begin{document}

\maketitle

\begin{abstract}
Diffusion models are an important tool for generative modelling, serving as effective priors in applications such as imaging and protein design. A key challenge in applying diffusion models for downstream tasks is efficiently sampling from resulting posterior distributions, which can be addressed using Doob's $h$-transform. This work introduces a self-supervised algorithm for fine-tuning diffusion models by learning the optimal control, enabling amortised conditional sampling. Our method iteratively refines the control using a synthetic dataset resampled with path-based importance weights. We demonstrate the effectiveness of this framework on class-conditional sampling, inverse problems and reward fine-tuning for text-to-image diffusion models. 
\end{abstract}



\section{Introduction}
Generative modelling aims to construct a sampling mechanism for an unknown data distribution $p_\mathrm{data}$ accessible only through a finite dataset.
Score-based diffusion models have recently become a particularly powerful framework for achieving this \cite{dhariwal2021diffusion,ho2020denoising,song2021Scorebased}.
They operate by gradually transforming the target distribution $p_{\mathrm{data}}$ through a drift $f_t$ and diffusion $\sigma_t$, obtained by simulating the stochastic differential equation (SDE)
\begin{align}\label{eq:forward_SDE}
    \dd \bX_t &= f_t(\bX_t) \,\dd t + \sigma_t \fwd{ \dd \rv{W}}_t, \quad \bX_0 \sim  p_{\mathrm{data}}.
\end{align}
We denote by $p_t$ the density of the solution $\bX_t$ at time $t$.
A widely used choice of coefficients is $f_t(\vx)=-\frac12 \vx$ and $\sigma_t=1$, known as the variance‑preserving SDE (VP‑SDE) \cite{song2021Scorebased}, for which $p_t$ converges exponentially fast to $\mathcal N(0,\mathbf{I})$.
To generate new samples from $p_\mathrm{data}$, diffusion models approximate the score function $s_t(\vx)=\nabla_\vx\ln p_t(\vx)$ with a neural network.
Given this approximation, they draw $x_T$ from $\mathcal N(0,\mathbf{I})\approx p_T$ and simulate the time‑reversed dynamics of \eqref{eq:forward_SDE}, often referred as reverse SDE, given by
\begin{align}
    \label{eq:back_sde}
    \dd \bX_t &= \left( f_t(\bX_t) - \sigma_t^2 s_t(\bX_t) \right) \,\dd t + \sigma_t \bwd{ \dd \rv{W}}_t,
\end{align}
where time evolves backward \cite{anderson1982reverse}.

In recent years, powerful foundation models such as Stable Diffusion \cite{esser2024scaling,rombach2022high} for images or RFDiffusion \cite{watson2023novo} for protein design have emerged. As these models provide strong general‑purpose generative capabilities, an important challenge is to adapt them effectively to specific downstream tasks.
Mathematically, this can be described as sampling from a \textit{tilted distribution} defined as  
\begin{align}
    \label{eq:tilted_dist}
    p_\text{tilted}(\vx) \propto p_\text{data}(\vx) \exp\left(\frac{r(\vx)}{\lambda} \right),
\end{align}
where $r\colon\R^n \to \R$ is a reward function specifying the task and $\lambda > 0$ is a temperature parameter.
To ensure that $p_\text{tilted}$ is a probability density function, we require that the normalising constant
\begin{equation}\label{eq:normalising_constant}
Z_r= \mathbb{E}_{\vx\sim p_\text{data}}[\exp(r(\vx)/\lambda)]
\end{equation}
is finite, which we assume throughout the paper.
The general formulation of the tilted distribution captures several applications, in particular:
\begin{itemize}
    \item[-] \textit{Bayesian inverse problems:} If $r(\vx)=\ln p(\vy|\vx)$ is chosen as the log-likelihood function in a Bayesian inverse problem, then $p_\mathrm{tilted}$ coincides with the posterior distribution $p(\vx|\vy)$ for the prior $\vx\sim p_\mathrm{data}$.
    \item[-] \textit{Class-conditional sampling:} If the samples from $p_\mathrm{data}$ can be divided into classes, we can generate samples belonging to class $\vc$ by choosing $r(\vx)=\ln p(\vc|\vx)$ as the log-probabilities of a classifier.
    \item[-] \textit{Fine-tuning based on human feedback:} We can improve the quality of generated samples based on human preferences by choosing $r(\vx)$ as a reward function learned from human feedback, see \cite{xu2024imagereward} for the training of such feedback functions.
\end{itemize}
To sample from $p_\mathrm{tilted}$, we modify the reverse SDE \eqref{eq:back_sde} by an additional control term $u_t$ as
\begin{align}\label{eq:controlled_SDE}
    \dd \bH_t &= \left( f_t(\bH_t) - \sigma_t^2 \left(s_t(\bH_t)+u_t(\bH_t)\right) \right) \,\dd t + \sigma_t \bwd{ \dd \rv{W}}_t.
\end{align}
Now, we aim to find an optimal control term $u_t^*$ such that for the initial condition $\bH_T$, approximately given by $\mathcal N(0,\mathbf{I})$, the solution $\bH_0$ of the controlled SDE \eqref{eq:controlled_SDE} at time $0$ follows the tilted distribution $p_\mathrm{tilted}$. One way to achieve this is to choose $u_t\coloneqq u_t^*=\nabla_{\vx} \ln h^r_t(\vx)$, where 
\begin{align}\label{eq:doob_control}
    u_t^*=\nabla_{\vx} \ln h^r_t(\vx),\quad \text{with} \quad h_t^r(\vx) = Z_r^{-1} \mathbb{E}\left[\exp(r(\bX_0)/\lambda)| \bX_t = \vx \right],
\end{align}
with $Z_r$ from \eqref{eq:normalising_constant}, which is referred as Doob's $h$-transform \cite{denker2024deft,rogers2000diffusions,vargas2023denoising}.
However, accessing the control $u_t^*$ from \eqref{eq:doob_control} directly is generally intractable. As a remedy, several indirect estimation methods for $u_t^*$ have been proposed recently, we refer to \cite{uehara2025reward} for an overview.

Inference-time methods, like classifier guidance \cite{dhariwal2021diffusion} and reconstruction guidance \cite{chung2022diffusion}, propose heuristic inexact approximations of $u_t^*$. While these methods do not require an additional training-phase, the approximation error leads to a bias, they often increase the computational cost for generating one sample and exhibit sensitivity to hyperparameters \cite{song2023solving}.

Here, we focus on post-training methods in which either $u_t$ is parametrised as a neural network or the weights of the score network for $s_t$ are modified so that it implicitly represents the sum of the original score function and the control term.
Learning the parameters of this additional or modified network requires an extra training stage. While some approaches rely on supervised objectives and therefore require additional task‑specific datasets \cite{denker2024deft,ruiz2022dreambooth,xu2024imagereward,zhang2023adding}, our interest lies in self-supervised post‑training methods that operate without additional data. 
Among those, online post-training methods directly optimise an objective defined by the reward function via reinforcement learning \cite{black2024training,clark2023directly,fan2024reinforcement,venkatraman2024amortizing} or stochastic optimal control \cite{denker2024deft,domingo2024adjoint,pidstrigach2025conditioning}. However, they often require path-wise gradient estimators such as REINFORCE \cite{williams1992simple} or policy gradients methods \cite{schulman2017proximal}, which may exhibit extensive computational costs or a high variance. 

An alternative to online reinforcement learning aligns generative models by fine-tuning them on their own high-reward outputs. Such methods typically follow an iterative ``sample $\rightarrow$ filter $\rightarrow$ fine-tune'' procedure, in which a reward function ranks model‑generated trajectories and the model is updated using only the top‑ranked samples. Prominent examples include REST \cite{gulcehre2023reinforced} and RAFT \cite{dong2023raft}, which have been applied to both large language models and diffusion models; see also \cite{de2005tutorial,fan2024reinforcement,lee2023aligning,peters2007reinforcement} for further references on this paradigm.
The method we introduce in this paper is closely related to this principle. However, unlike most existing approaches, we can directly characterise the limit of the iterative procedure and show that it corresponds to the tilted distribution \eqref{eq:tilted_dist}.

\subsection{Contributions}
We propose a self-supervised framework for estimating the control without access to samples from the tilted distribution. 
Starting with an initial estimate of the control $u_t$, our method iterates the following three steps:
\begin{itemize}
    \item[1.] First, we \emph{sample a batch of trajectories from the diffusion model} with our current estimate of the control.
    \item[2.] Next, we use path-based importance weights to \emph{filter the dataset} by rejecting samples which do not fit to the tilted distribution. Consequently, the filtered dataset describes the filtered distribution better than the original one. The corresponding importance weights can be computed on the fly during sampling and do not cause additional costs.
    \item[3.] Finally, we use a supervised fine-tuning loss \cite{denker2024deft} to \emph{update our estimation of the control}.
\end{itemize}
From a theoretical viewpoint, we prove in Theorem~\ref{the:} that our iterative importance fine-tuning decreases a certain stochastic optimal control loss function in each iteration. 
We apply our method for class conditional sampling on MNIST, super-resolution, and reward-based fine-tuning of Stable Diffusion. We emphasise that our importance-based fine-tuning  \emph{does not require to differentiate the score-function of the base-model}. Since we fine-tune using the score matching objective, we never have to backpropagate through the generation process.  
In particular, it can be applied for very large base-models where other methods run out of memory. 

\subsection{Outline}

The rest of the paper is structured as follows. In Section \ref{sec:background} we give the necessary background on diffusion models and supervised fine-tuning. The path-based importance weights and the resampling step are presented in Section \ref{sec:selfsupervised}. We present experiments on class conditional sampling for MNIST, super-resolution, and reward-based fine-tuning of Stable Diffusion \cite{rombach2022high} in Section~\ref{sec:experiments}. Conclusions are drawn in Section~\ref{sec:concl}.

\subsection{Notation}
Let $\mathcal{P}(\Omega)$ be the space of probability measures on $C([0,T], \mathbb{R}^n)$.
Throughout this paper, we will denote by $\sP^\mathrm{data}\in \mathcal{P}(\Omega)$ and $\sP^\mathrm{tilted}\in \mathcal{P}(\Omega)$ the path measure of the solution $\bX_t$ in \eqref{eq:forward_SDE} with initial conditions $\bX_0\sim p_\mathrm{data}$ and $\bX_0\sim p_\mathrm{tilted}$.
We also denote by $p_t^\mathrm{tilted}$ the marginal of $\sP^\mathrm{tilted}$ at time $t$. 
For a control function $u_t$, we denote by $\sP^u$ the path measure of the controlled reverse SDE \eqref{eq:controlled_SDE} with initial condition $\bH_T\sim p_T^\mathrm{tilted}$.
In addition, we denote for $t_1\geq t_2$ by $\fwd{p}_{t_1|t_2}(\vx_{t_1}|\vx_{t_2})$ the conditional densities of $\bX_{t_1}$ given $\bX_{t_2}$ in the forward SDE \eqref{eq:forward_SDE}. Note that solutions of SDEs are Markov processes, such that the conditional densities $\fwd{p}_{t_1|t_2}(\vx_{t_1}|\vx_{t_2})$ do not depend on the initial distribution of $\bX_0$.
On the other hand, the conditional densities for $t_1<t_2$ depend on the initial distribution of $\bX_0$. For $\bX_0\sim p_\mathrm{data}$ and $\bX_0\sim p_\mathrm{tilted}$ we denote them by $\bwd{p}_{t_1|t_2}^\text{data}(\vx_{t_1}|\vx_{t_2})$.
Further, $\vx_{[0,T]}$ denotes a trajectory $(\vx_t)_{t\in[0,T]}$.

\section{Tilted Score and Variational Formulation of Doob's $h$-transform}
\label{sec:background}

In this section, we will examine the relation of Doob's $h$-transform to the score of the tilted distribution and its formulation as minimiser of certain loss functions. 
We start with computing the Radon-Nikodym derivative between two path measures generated by the same SDE with different initial distributions. The proof is a short computation, given below for convenience.

\begin{lemma} \label{lemma:sol_sde}
Assume that $\sP$ and $\sQ$ the path measures of the solution for the same SDE $\dd \bY_t = f_t(\bY_t) \,\dd t + \sigma_t \fwd{ \dd \rv{W}}_t$ with initial conditions $\bY_0 \sim p$ and $\bY_0 \sim q$, respectively. We have
\begin{align}
    \frac{p(\vx_0)}{q(\vx_0)} = \frac{\dd \sP}{\dd \sQ}(\vx_{[0,T]}). 
\end{align}
\end{lemma}
\begin{proof}
Denote by $\sP=p\times \sP_{\vx_0}$ and $\sQ=q \times \sQ_{\vx_0}$ the disintegrations of $\sP$ and $\sQ$ with respect to time $0$. Since $\sP$ and $\sQ$ are both path measures for solutions of the SDE  $\dd \bY_t = f_t(\bY_t) \,\dd t + \sigma_t \fwd{ \dd \rv{W}}_t$ we have that for any $\vx_0$ that $\sP_{\vx_0}$ and $\sQ_{\vx_0}$ coincide with the path measure of this SDE with initial condition $\bY_0 \sim \delta_{\vx_0}$. In particular, the disintegrations $\sP_{\vx_0}$ and $\sQ_{\vx_0}$ coincide. Hence, we can conclude 
\begin{align}
    \frac{\dd \sP}{\dd \sQ}(\vx_{[0,T]})  =\frac{p(\vx_0)}{q(\vx_0)} \frac{\dd \sP_{\vx_0}}{\dd \sQ_{\vx_0}}(\vx_{[0,T]}) = \frac{p(\vx_0)}{q(\vx_0)}.
\end{align}
This completes the proof.
\end{proof}

As a consequence of the lemma we obtain that the Radon-Nidodym derivative of $\sP^\mathrm{tilted}$ and $\sP^\mathrm{data}$ is given by the $h$-transform at time zero. More precisely, we get
\begin{equation}
    \label{eq:optimal_path_measure}
    \frac{\dd \sP^\mathrm{tilted}}{\dd \sP^\mathrm{data}}(\vx_{[0,T]}) =  h_0^r(\vx_0) = \frac{1}{Z_r} \exp(r(\vx_0)/\lambda).
\end{equation}
Using this formula, we can now represent the tilted path measure as the minimiser of the so-called free energy functional, which is known as the Donsker-Varadhan variational formula~\cite{donsker1976asymptotic}. The proof is again a short computation which we assemble below from the literature, see for example \cite[Chapter 6]{rogers2000diffusions}. 

\begin{theorem}[Variational Principle for Tilted Path Measures]
\label{th:variational_principle}
The tilded path measure $\sP^\mathrm{tilted}$ is the unique minimiser of the
free energy $\mathcal{F}: \mathcal{P}(\Omega) \to \mathbb{R} \cup \{+\infty\}$ defined by
\begin{equation}\label{eq:free_energy}
    \mathcal{F}(\sP) = \KL(\sP \| \sP^\mathrm{data}) -\frac{1}{\lambda} \mathbb{E}_{\vx_{[0,T]} \sim \sP}[r(\vx_0)].
\end{equation}
\end{theorem}
\begin{proof}
Taking the logarithm in \eqref{eq:optimal_path_measure} on both sides, we obtain for any $x_{[0:T]}$ that $r(\vx_0)/\lambda=\ln \frac{\dd \sP^\mathrm{tilted}}{\dd \sP^\mathrm{data}}(\vx_{[0,T]})+ \ln Z_r$.
Inserting this into the free energy functional we get
\begin{align}
    \mathcal F(\sP)&=\E_{x_{[0:T]}\sim\P}\left[\ln \frac{\dd \sP}{\dd \sP^\mathrm{data}}(x_{[0:T]})-\ln \frac{\dd \sP^\mathrm{tilted}}{\dd \sP^\mathrm{data}}(\vx_{[0,T]})\right]-\ln Z_r\\
    &=\E_{x_{[0:T]}\sim\P}\left[\ln \frac{\dd \sP}{\dd \sP^\mathrm{tilted}}(x_{[0:T]})\right]-\ln Z_r=\mathrm{KL}(\sP\|\sP^\mathrm{tilted})-\ln Z_r.
\end{align}
Now the claim follows from the properties of the Kullback-Leibler divergence.
\end{proof}

Next, we represent the tilted scores $\nabla_{\vx}\log p_t^\mathrm{tilted}(\vx)$ by the data score $s_t(\vx)=\nabla_{\vx}\ln p_t(\vx)$ and the $h$-transform.
The following proposition was proved in \cite[Thm 3.1, Part 2)]{denker2024deft} for the case that $r$ is the log-likelihood function in a Bayesian inverse problem. We state it for general reward functions and give a proof using similar computations as in \cite{denker2024deft}.
A direct consequence of this decomposition is that the reverse SDE \eqref{eq:controlled_SDE} controlled with $u_t^*(\vx)=\nabla_{\vx}\ln h_t^r(\vx)$ indeed generates the tilted distribution.

\begin{proposition}[Score Decomposition]
\label{lem:score_decomp}
The tilted score decomposes into the unconditional score $s_t(\vx)$ and the gradient of the logarithm of the $h$-transform:
\begin{align}
    \nabla_\vx \ln p_t^\mathrm{tilted}(\vx) = s_t(\vx) + \nabla_\vx \ln h_t^r(\vx).
\end{align}
In particular, we obtain for $u_t^*(\vx)=\nabla_{\vx}\ln h_t^r(\vx)$ that the solution of the controlled SDE \eqref{eq:controlled_SDE} with initial condition $\bH_T\sim p_T^\mathrm{tilted}$ fulfils $\bH_0\sim p_\mathrm{tilted}$.
\end{proposition}
\begin{proof}
Using the computation rules for conditional probabilities we have
\begin{align}
    h_t^r(\vx_t) &= \frac{1}{Z_r} \int \exp(r(\vx_0)) \bwd{p}_{0|t}^\text{data}(\vx_0 | \vx_t) \dd \vx_0\\
    &= \frac{1}{Z_r\,p_t(\vx_t)} \int \exp(r(\vx_0)) p_\text{data}(\vx_0) \fwd{p}_{t|0}(\vx_t|\vx_0) \dd \vx_0 \\ 
    &= \frac{1}{p_t(\vx_t)} \int p_\text{tilted}(\vx_0)  \fwd{p}_{t|0}(\vx_t|\vx_0) \dd \vx_0 = \frac{p_t^\text{tilted}(\vx_t)}{p_t(\vx_t)}.
\end{align}
Taking the logarithm and gradient on both sides yields the first claim. The second claim directly follows from the fact that \eqref{eq:controlled_SDE} is now the time reversal of the forward SDE \eqref{eq:forward_SDE} with initial condition $\bX_0\sim p_\mathrm{tilted}$.
\end{proof}

In order to generate sample from $p_\mathrm{tilted}$, we have to initialise the reverse SDE \eqref{eq:controlled_SDE} controlled by $u_t^*$ with $\bH_T=p^\mathrm{tilted}_T$ instead of $p_T$. 
However, the authors of \cite[Proposition G.2]{denker2024deft} show that also $p_T^\mathrm{tilted}$ converges exponentially fast to $\mathcal N(0,\mathbf{I})$ for the commonly used forward SDEs.
In particular, we may approximate $p_T^\mathrm{tilted}$ by $\mathcal N(0,\mathbf{I})$, in direct analogy with the uncontrolled setting.

To compute the optimal control $u_t^*$ numerically, the authors of \cite{denker2024deft} proposed two kinds of loss functions, which are summarized in the following theorem. The first part directly follows from Proposition~\ref{lem:score_decomp}. For the second part, we refer to \cite[Thm 3.1]{denker2024deft}.

\begin{theorem}[Loss functions for Computing $u_t^*$]\label{th:deft}
The optimal control $u_t^*(\vx)=\nabla_\vx\ln h_t^r(\vx)$ is the unique minimiser of the score-matching (SM) loss $\mathcal L_{SM}$ and the stochastic control (SC) loss $\mathcal L_{SC}$ given by
\begin{align}
\label{eq:supervised_deft}
    \mathcal L_{SM}(u_t)&= \mathop{\mathbb{E}}_{\substack{ \vx_0 \sim p_\mathrm{tilted} \\ t\sim \mathrm{U}(0,T), \vx_t \sim \fwd{p}_{t|0}(\cdot|\vx_0)}  }\left[ \left\| \left({u_t(\vx_t)}\!+\!s_t(\vx_t) \right)\!-\!\nabla_{\vx_t}\ln \fwd{p}_{t|0}(\vx_t  |\vx_0) \right\| ^2\right],\\
    \label{eq:control_deft}
    \mathcal L_{SC}(u_t) &= \mathbb{E}_{\vx_{[0:T]}\sim\sP^u}\left[\frac12\int_0^T\sigma_t^2\|u_t(\vx_t)\|^2\d t-r(\vx_0)/\lambda\right].
\end{align}
Moreover, the SC loss is related to the free energy from \eqref{eq:free_energy} by $\mathcal L_{SC}(u_t)=\mathcal F(\P^u)+\mathrm{const}$.
\end{theorem}

The two loss functions from the theorem differ in the computational requirements and assumptions on availability on the dataset: 
The score-matching loss $\mathcal L_{SM}$ is a simple mean squared error which only needs to consider one time point $t$ at once. However, its computation requires access to samples from $p_\mathrm{tilted}$, which are often not available.
In contrast, the stochastic control loss $\mathcal L_{SC}$ is unsupervised and does not require additional data. However, it requires samples from the reverse SDE \eqref{eq:controlled_SDE} controlled by the argument $u_t$ of the loss function. Such loss functions are often referred as ``simulation-based'' and are computationally very costly, since gradients must be backpropagated through the entire trajectory. 
This typically requires path-wise estimators like REINFORCE \cite{williams1992simple,fan2024reinforcement} or policy gradient methods \cite{schulman2017proximal}, which often suffer from high variance. Techniques such as VarGrad \cite{richter2020vargrad} or trajectory balance \cite{malkin2022trajectory} can reduce computational cost, but memory usage still grows linearly with trajectory length.

\section{Self-Supervised Importance Fine-tuning}
\label{sec:selfsupervised}
In this section, we propose self-supervised importance fine-tuning, which leverages the computational efficient score-matching loss \eqref{eq:supervised_deft} without requiring samples from $p_\mathrm{tilted}$.
To this end, we combine the recently proposed rejection sampling steps from \cite{hertrich2024importance} to generate a synthetic dataset which replaces $p_\mathrm{tilted}$ within \eqref{eq:supervised_deft}. Additionally, we prove that this defines a descent algorithm for the free energy \eqref{eq:free_energy}.

\subsection{Method Description} \label{sec:algorithm}
Starting with an initial estimate $u^{0}$ of $u^*$, for instance $u_t^0(\vx)=0$ for all $\vx$, we iteratively estimate a sequence $u^k$, $k=1,2,...$ of control terms by the fixed-point iteration consisting out of the following three steps:
\begin{enumerate}
    \item Sample trajectories $\vx_{[0:T]}^{(n)}$ from $\sP^{u^{k}}$ for $n=1,...,N$.
    \item Compute acceptance probabilities $\alpha_{k}(\vx_{[0:T]}^{(n)})$ for each trajectory $\vx_{[0:T]}^{(n)}$, which we will derive in \eqref{eq:acceptance_prob} and \eqref{eq:acceptance_prob_explicit} of Subsection~\ref{sec:importance}. We denote the distribution of $\{\vx_0:\vx_{[0:T]}\text{ is accepted}\}$ by $\tilde p_{u^k}$. In practice, $\tilde p_{u^k}$ is represented by a finite dataset which contains $\vx_0^{(n)}$ with probability $\alpha_{k}(\vx_{[0:T]}^{(n)})$ for $n=1,...,N$.
    \item We estimate the control $u^{k+1}$ by minimizing the fine-tuning loss
    \begin{align}\label{eq:supervised_finetuning_loss}
\mathcal L_{FT}(u_t)=\mathop{\mathbb{E}}_{\substack{ \vx_0 \sim \tilde p_{u^k} \\ t\sim \mathrm{U}(0,T), \vx_t \sim \fwd{p}_{t|0}(\cdot|\vx_0)}  }\left[ \left\| \left({u_t(\vx_t)}\!+\!s_t(\vx_t) \right)\!-\!\nabla_{\vx_t}\ln \fwd{p}_{t|0}(\vx_t  |\vx_0) \right\| ^2\right].  
\end{align}
    It coincides with the score-matching loss $\mathcal L_{SM}$ from \eqref{eq:supervised_deft} with the only difference that the (unknwon) distribution $p_\mathrm{tilted}$ is replaced by the (known) distribution $\tilde p_{u^k}$.
\end{enumerate}

The rest of this section is organised as follows. First, we derive in Subsection~\ref{sec:importance} the acceptance probabilities $\alpha^{u^{k}}(\vx_{[0:T]}^{(n)})$ from step 2 and prove that the free energy $\mathcal F(\sP^{u^k})$, or equivalently the stochastic control loss $\mathcal L_{SC}(u^k)$, decreases over the iterations.
Finally, in the context of large text-to-image models, various additional regularisation strategies and approximations are often employed to improve computational efficiency. We will discuss these in more detail in Subsection~\ref{sec:training_network}.

\subsection{Importance-based Acceptance Probabilities and Convergence}\label{sec:importance}

To define the acceptance probabilities in the second stage of our algorithm, we rely on importance sampling. Instead of drawing directly from $p_\mathrm{tilted}$, we sample from a more convenient proposal distribution $q$ and assign each candidate point $\vx_0$ the weight $\frac{p_\mathrm{tilted}(\vx_0)}{q(\vx_0)}$. With these weights, the resulting weighted samples are distributed according to $p_\mathrm{tilted}$.
Here, we choose the proposal distribution $q$ as the distribution $p_{u^k}$ of the solution $\bH_0$ of the reverse SDE \eqref{eq:controlled_SDE} with control $u^k$ at time $0$. 
Using the factorisation $p_\text{tilted}(\vx_0)\propto p_\text{data}(\vx_0)\exp(r(\vx_0)/\lambda)$, we can rewrite the importance weights as
\begin{align}
\frac{p_\mathrm{tilted}(\vx_0)}{p_{u^k}(\vx_0)}
= \frac{p_\text{tilted}(\vx_0)}{p_\text{data}(\vx_0)} \cdot \frac{p_\text{data}(\vx_0)}{p_{u^k}(\vx_0)}
\propto \exp(r(\vx_0)/\lambda) \frac{p_\text{data}(\vx_0)}{p_{u^k}(\vx_0)}.
\end{align}
Unfortunately, computing the marginal densities $p_\text{data}$ and $p_{u^k}$ is generally intractable. However, using Lemma~\ref{lemma:sol_sde}, we can represent their ratio by 

\begin{corollary}
Let $(u^k)_k$ be generated by the algorithm in Section~\ref{sec:algorithm}. Then, we have for $k\geq 1$ and any $x_{[0:T]}$ that
\begin{equation}
\frac{p_\mathrm{data}(\vx_0)}{p_{u^k}(\vx_0)}=\frac{\dd \sP^\mathrm{data}}{\dd \sP^{u^k}}(\vx_{[0:T]}).
\end{equation}
\end{corollary}
\begin{proof}
By the definition from step 3 in Section~\ref{sec:algorithm}, we know that for $k\geq 1$ the control $s_t+u_t^k$ minimizes the score matching loss for the distribution $\tilde p_{u^k}$. In particular, this implies that $\sP^{u^k}$ is the path measure of the forward SDE \eqref{eq:forward_SDE} with initial condition $\bX_0\sim \tilde p_{u^k}$. Hence, we obtain by Lemma~\ref{lemma:sol_sde} that the density ratio $\frac{p_\text{data}(\vx_0)}{p_{u^k}(\vx_0)}$ coincides with the Radon-Nikodym derivative $\frac{\sP^\text{data}}{\sP^{u^k}}(\vx_{[0:T]})$ for any trajectory $\vx_{[0:T]}$. 
\end{proof}

Next, we compute the Radon–Nikodym derivative (RND) from the lemma explicitly using the framework of \cite{nusken2024transport}, leading to the following result. The derivations are included in Appendix~\ref{app:RND_SDE}, both in continuous and in discrete time.

\begin{lemma} \label{lemma:rnd}
For any control $u$ with corresponding path measure $\sP^u$ of the controlled SDE \eqref{eq:controlled_SDE} and any continuous trajectory $x_{[0:T]}$, we have 
\begin{align}\label{eq:rnd}
   \frac{\sP^\mathrm{data}(\vx_0)}{\sP^{u}(\vx_0)}
\propto \exp \left( \frac{1}{\lambda} r(\vx_0) - \frac12 \int_0^T \sigma_t^2\| u_t(\vx_t) \|_2^2  dt + \int_0^T \sigma_t u_t(\vx_t)^\top dW_t\right).
\end{align}
\end{lemma}

\begin{remark}[Computational Tractability]
These path-wise importance weights can be computed concurrently with sampling without any computational overhead, using the same time-discretiation for the integrals as in the SDE solver. For numerical stability, we perform these computations in the log-space. 
\end{remark}

In practice, importance weights for high-dimensional distributions are usually imbalanced, leading to high variances and low effective sample sizes. In the context of rejection-based algorithms, this leads to very small acceptance rates.
As a remedy, \cite{hertrich2024importance} proposed to relax the importance weights by considering the acceptance probabilities defined as
\begin{align}\label{eq:acceptance_prob}
\alpha_{k}(\vx_{[0:T]})&=\min\left(1,\frac{\dd \sP^\text{data}}{\dd \sP^u}(\vx_{[0,T]}) \frac{\exp(r(\vx_0)/\lambda)}{c}\right),
\end{align}
where $c$ is a hyper-parameter, which controls the acceptance ratio by the following remark.

\begin{remark}[Choice of $c$]
Based on the observation that $\alpha(\vx_{[0:T]})$ is decreasing in $c$, \cite[Rem 4.4]{hertrich2024importance} suggest to find $c$ such that $\E[\alpha(\vx_{[0:T]})]=\rho$ for some prescribed acceptance rate $\rho$. Such a $c$ can easily be found by some bisection search.
In this way it is always ensured that a sufficient amount of samples is accepted, which circumvents one of the major drawbacks from rejection sampling with importance weights.
In our numerical examples, we choose the acceptance ratio $\rho$ for each experiment separately between $0.1$ and $0.6$.
\end{remark}

Inserting the formula derived in \eqref{eq:rnd} into \eqref{eq:acceptance_prob}, we obtain the explicit representation of the acceptance probabilities
\begin{align}\label{eq:acceptance_prob_explicit}
\alpha_{k}(\vx_{[0:T]})\!=\!\min\left(1,\frac{1}{\tilde c}\exp \left( r(\vx_0)/\lambda\!- \frac12 \int_0^T\!\sigma_t^2\| u_t(\vx_t) \|_2^2  dt + \int_0^T\!\sigma_t u_t(\vx_t)^\top dW_t\right)\right),
\end{align}
where $\tilde c$ is a hyper-parameter combining the parameter $c$ from \eqref{eq:acceptance_prob} and the normalising constant from \eqref{eq:rnd}.
Finally, we prove that the algorithm proposed in Subsection~\ref{sec:algorithm} always reduces the free energy \eqref{eq:free_energy} over the iterations, i.e., that $\mathcal F(\sP^{u^{k+1}})\leq \mathcal F(\sP^{u^{k+1}})$.
We include the proof in Appendix \ref{app:proof_resampling}.
\begin{theorem}\label{the:}
Let $\alpha_k(\vx_{[0:T]})$ be the acceptance probability from \eqref{eq:acceptance_prob} for some control $u^k$ and let $\vx_{[0,T]}$ be a trajectory. Further, we denote by $\tilde \sP^{u^k}$ the distribution of accepted paths and by $\tilde p_{u^k}$ its marginal at time $1$ and by $u_t^{k+1}\in\argmin_{g_t}\mathcal L_{FT}(g_t)$ with $\mathcal L_{FT}$ from \eqref{eq:supervised_finetuning_loss}.
Then it holds for the free energy $\mathcal F$ from  \eqref{eq:free_energy} that
\begin{equation}
\label{eq:the_claim}
\mathcal F(\sP^{u^{k+1}})\leq\mathcal F(\tilde{\sP}^{u^k})\leq\mathcal F(\sP^{u^k}).
\end{equation}
\end{theorem}

\subsection{Training and Network Parametrisation}
\label{sec:training_network}

In our experimental section, we will consider examples involving large base models that approximate the data distribution $p_\mathrm{data}$. For example, Stable Diffusion \cite{rombach2022high} has 800 million parameters, which makes generating new samples and computing the score along with its derivative computationally very expensive.  
Moreover, iterative retraining of generative models on synthetic data has been shown to lead to performance degradation, including phenomena such as mode collapse \cite{alemohammad2023self,bertrand2023stability,shumailov2023curse}.
To fine-tune models of this scale, we rely on several additional implementation techniques that have previously been introduced in the literature. More precisely, we use replay buffers, implicit biases in the network parametrisation and an additional regularisation term within the fine-tuning loss $\mathcal L_{FT}$. In the following, we describe these techniques more in detail.

\paragraph{Replay buffer} Sampling from the current model is the most computationally expensive part of the algorithm. Inspired by \cite{midgley2022flow,sendera2024improved}, we employ an un-prioritised replay buffer of fixed size. We store the accepted samples from the current model and add them to the buffer. Once the buffer reaches its maximum capacity, a portion of samples is removed.
The samples selected for removal from the buffer can either be chosen randomly or as the oldest entries. In the random selection scenario, the replay buffer acts similarly to a moving average method, and a similar conclusion to Theorem~\ref{the:} is applicable. Nevertheless, in practice, we choose to remove the oldest samples, which produces comparable results and achieves faster convergence. Additional details are provided in Appendix~\ref{app:replay}.
To minimise the fine-tuning loss~\eqref{eq:supervised_finetuning_loss}, we randomly draw batches from the buffer.

\paragraph{Network parametrisation} The parametrisation of the control $u_t$ has a crucial effect on performance and convergence speed \cite{he2025no}. Motivated by the network parametrisation in sampling applications with diffusion \cite{vargas2023denoising,zhang2021path} and fine-tuning approaches \cite{denker2024deft,venkatraman2024amortizing}, we make use of a \textit{reward-informed inductive bias} given as 
\begin{align}
\label{eq:lkhd_informed_model}
    u^\theta_t(\vx_t) = \text{NN}_1(\vx_t, t) + \text{NN}_2(t) \nabla_{\hat \vx_0} r(\hat \vx_0(\vx_t)),
\end{align}
where $\hat \vx_0(\vx_t) \approx \mathbb{E}[\vx_0|\vx_t]$ is the denoised Tweedie estimate given the pre-trained unconditional diffusion model, $\text{NN}_1:\R^d \times [0,T] \to \R^d$ is a vector-valued and $\text{NN}_2:[0,T] \to \R^d$ a scalar-valued neural network. We initialise the last layer of $\text{NN}_1$ to be zero and $\text{NN}_2$ to be constant. However, for text-to-image diffusion models we instead make use of prior work, see e.g. \cite{venkatraman2024amortizing, fan2024reinforcement}, and use the parameter efficient LoRA method \cite{hu2021lora}. 

\paragraph{KL Regularisation} Similar to \cite{fan2024reinforcement}, we found that using a KL regulariser was useful to further improve diversity in the self-supervised setting. The KL divergence between $p_u$ and $p_\text{data}$ can be bounded as 
\begin{align}
    \label{eq:kl_reg}
    D_\text{KL}(p_u,p_\text{data}) \le D_\text{KL}(\sP^u,\sP^\text{data}) &=\mathbb{E}_{\bH_t \sim \sP^u}\left[ \int_0^T\!\sigma_t^2 \| u_t(\bH_t) \|_2^2 d t \right] \\ &= \mathbb{E}_{\bH_t \sim \sP_t^u, t \sim U(0,T)}\left[\sigma_t^2 \| u_t(\bH_t) \|_2^2 \right], 
\end{align}
i.e., the norm of the control over trajectories, see Appendix \ref{app:RND_SDE} for a derivation. Thus, our regularised version of the fine-tuning objective \eqref{eq:supervised_finetuning_loss} reads 
\begin{align}
    \mathcal{L}(u) = \mathop{\mathbb{E}}_{\substack{ \vx_0 \sim \tilde p_{u^k} \\ t\sim \mathrm{U}(0,T), \vx_t \sim \fwd{p}_{t|0}(\cdot|\vx_0)}  }\left[ \left\| \left({u_t(\vx_t)}\!+\!s_t(\vx_t) \right)\!-\!\nabla_{\vx_t}\ln \fwd{p}_{t|0}(\vx_t  |\vx_0) \right\|^2 + \alpha_\text{KL} \| u_t(\vx_t) \|^2  \right].  
\end{align}

\section{Experiments}
\label{sec:experiments}
In this section, we present a toy example on a 2D dataset, class conditional sampling for MNIST, super-resolution, and finally results for reward fine-tuning of text-to-image diffusion models. The code is available\footnote{\url{https://github.com/alexdenker/IterativeImportanceFinetuning}}.

\begin{figure}[htbp]
    \centering
    \begin{subfigure}{0.15\textwidth}
        \centering
        \includegraphics[width=\linewidth]{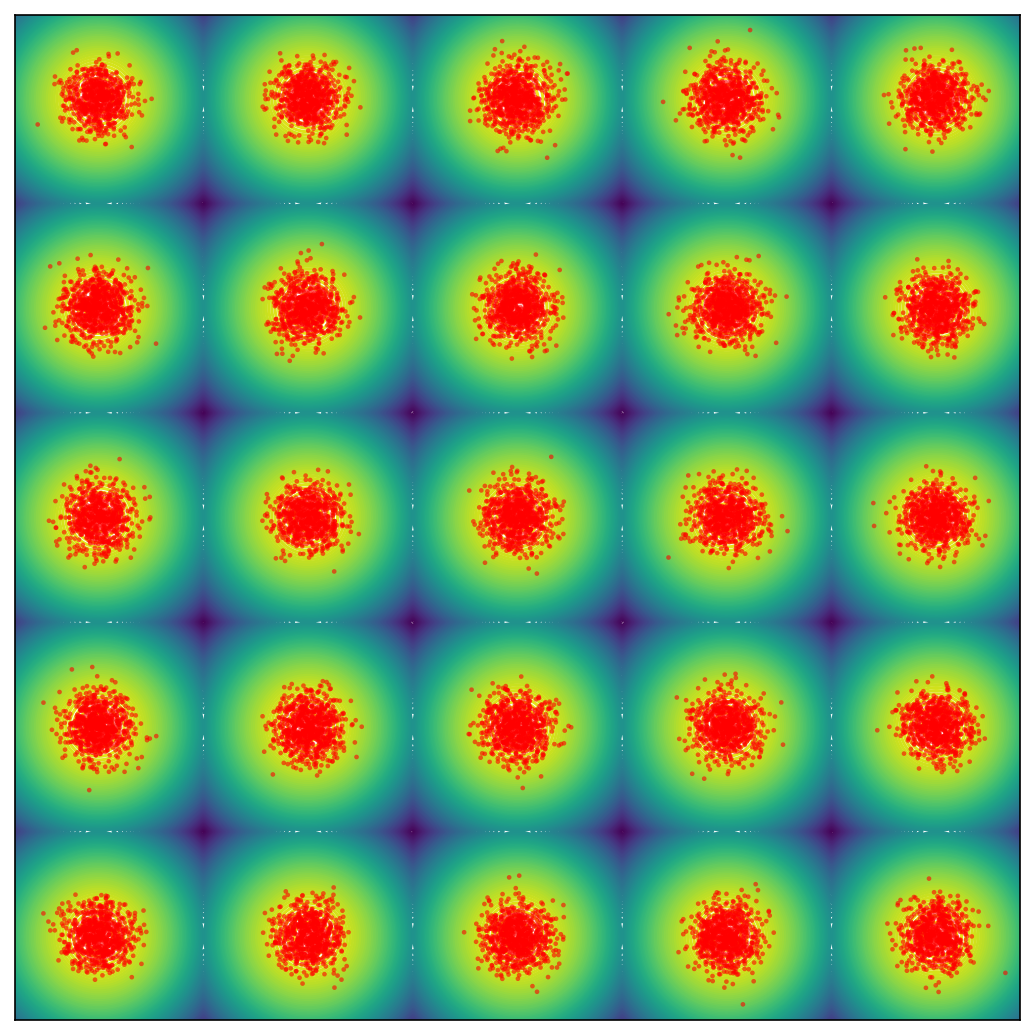}
        \caption*{Prior}
    \end{subfigure}
    \begin{subfigure}{0.15\textwidth}
        \centering
        \includegraphics[width=\linewidth]{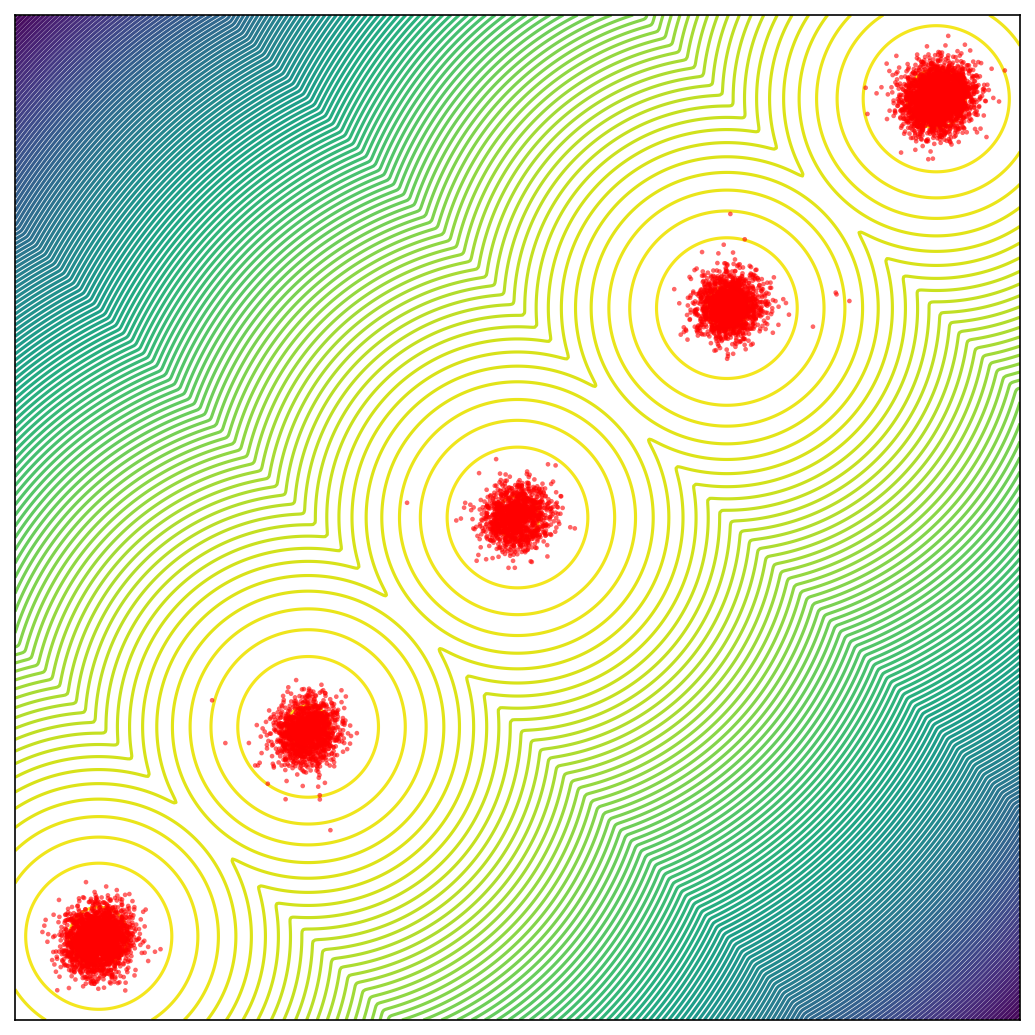}
        \caption*{Diagonal}
    \end{subfigure}
    \begin{subfigure}{0.15\textwidth}
        \centering
        \includegraphics[width=\linewidth]{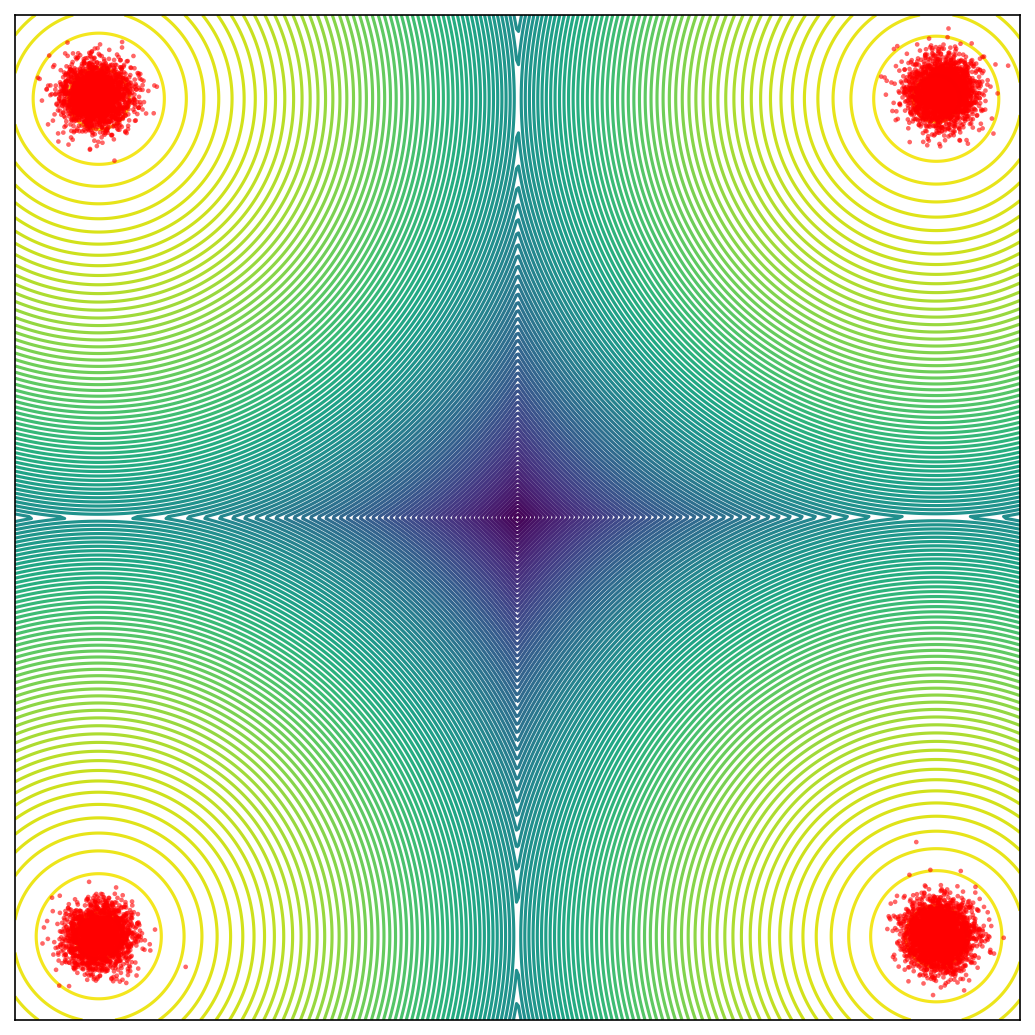}
        \caption*{Corners}
    \end{subfigure}
    \begin{subfigure}{0.15\textwidth}
        \centering
        \includegraphics[width=\linewidth]{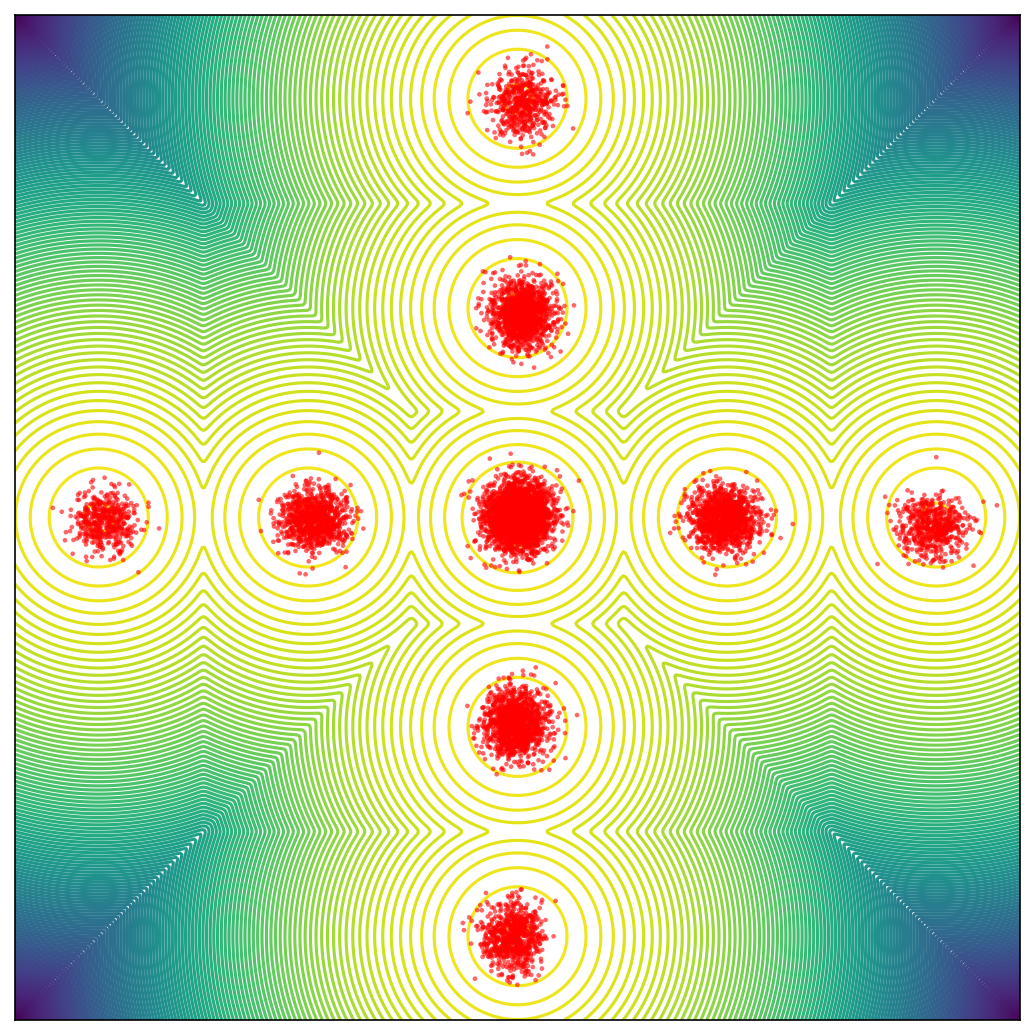}
        \caption*{Cross}
    \end{subfigure}
    \begin{subfigure}{0.15\textwidth}
        \centering
        \includegraphics[width=\linewidth]{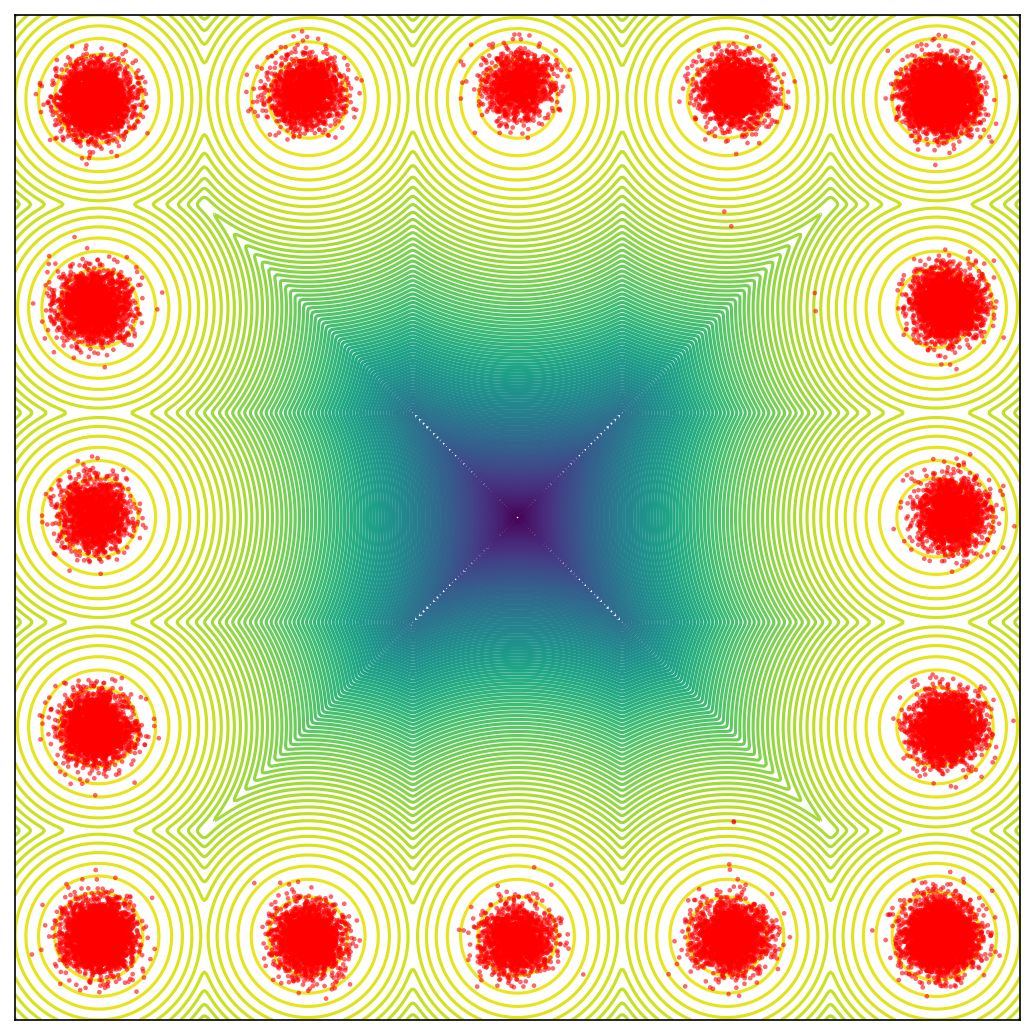}
        \caption*{Ring}
    \end{subfigure}
    \begin{subfigure}{0.15\textwidth}
        \centering
        \includegraphics[width=\linewidth]{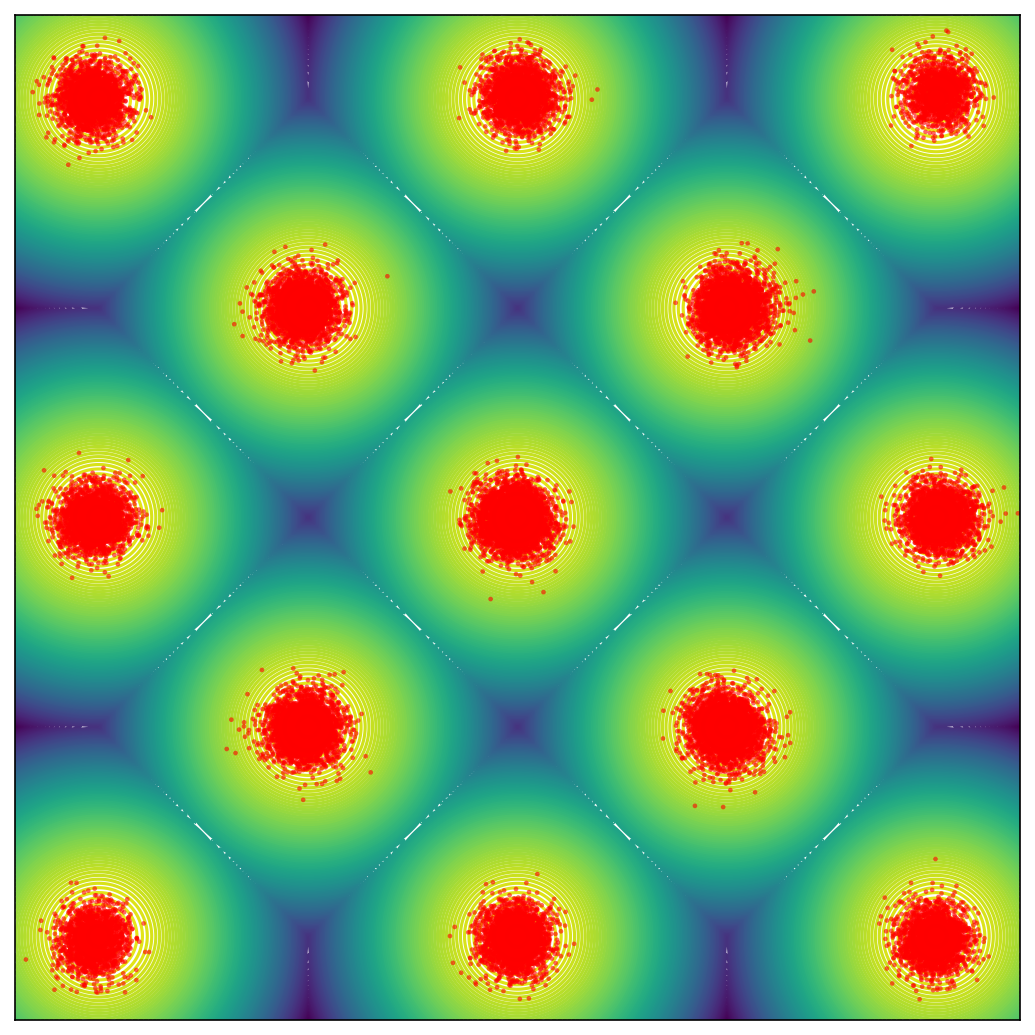}
        \caption*{Checkerboard}
    \end{subfigure}
    \caption{GMM prior and several posterior configurations: Diagonal, Corners, Cross, Ring and Checkerboard.}
    \label{fig:gmm_targets}
\end{figure}

\subsection{2D Toy example}
\label{sec:2Dtoyexample}
\begin{figure}
    \centering
    \includegraphics[width=1.0\linewidth]{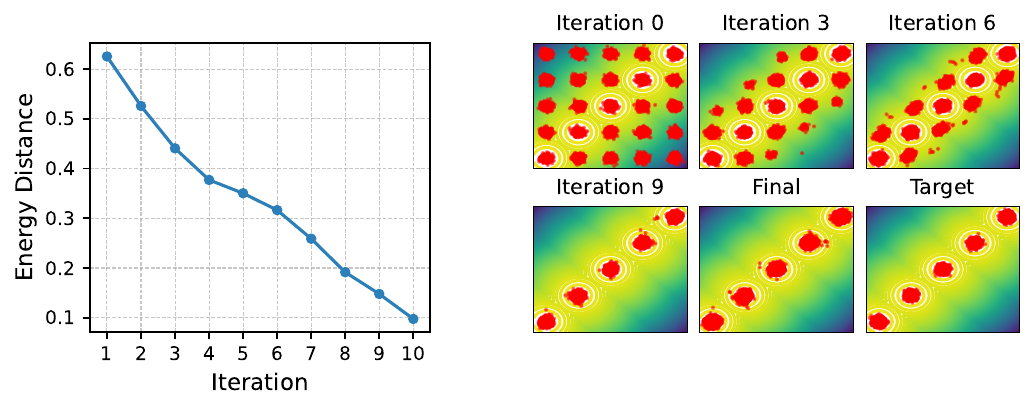}
    \caption{Evolution of the fine-tuned model across outer iterations. Left: energy distance as a function of iteration.}
    \label{fig:gmm_progression}
\end{figure}

We validate our method on a two-dimensional toy problem for which the exact target distribution is analytically available. 
The prior distribution $p_\text{data}(\vx)$ is defined as a Gaussian Mixture Model (GMM) with 25 equally weighted modes arranged on a grid. All mixtures are weighted equally with means in $\{-2.5, -1.25, 0, 1.25, 2.5\}^2$ and covariance $\Sigma=0.1 \mathbf{I}$. 
Our goal is to sample from the tilted distribution in \eqref{eq:tilted_dist}. 
We define a reward function as the log-likelihood ratio between a target distribution $p_{\text{target}}(\vx)$ and the prior, $r(\vx) = \log p_{\text{target}}(\vx) - \log p_{\text{data}}(\vx)$, so that the tilted distribution coincides exactly with the target, i.e., $p_\text{tilted}(\vx) = p_\text{target}(\vx)$ for $\lambda=1.0$.
Since the score functions for GMMs are available in closed form, we can apply our approach without introducing approximation errors in the score. 
We consider several choices of target distributions $p_\text{target}$, each defined as a GMM that reweights or selects subsets of the prior modes:
\begin{enumerate}[nosep,leftmargin=0.5em]
    \item[] \textbf{Diagonal}: five modes along the main diagonal from bottom-left to top-right, with increased weight on the two extreme diagonal modes.
    \item[] \textbf{Corners}: the four corner modes of the grid only, with asymmetric weights to create two diagonally opposite corners.
    \item[] \textbf{Cross}: a cross-shaped configuration consisting of the central row and central column, with the center mode assigned a higher weight.
    \item[] \textbf{Ring}: the perimeter of the grid, with increased weight on the four corner modes.
    \item[] \textbf{Checkerboard}: an alternating checkerboard pattern over the full grid, retaining every other mode with equal weights.
\end{enumerate}
We report results for all target configurations in Figure~\ref{fig:gmm_targets}. For each setting, we visualise samples from the final fine-tuned model alongside the log-density of the corresponding target GMM. Across all targets, the learned distributions closely match the intended mode locations. Figure~\ref{fig:gmm_progression} illustrates the dynamics of the fine-tuning process for the diagonal target. Starting from the prior distribution, probability mass is progressively shifted toward the desired diagonal modes while non-target modes are attenuated. This transition is smooth, indicating a stable optimisation process. 
To quantify convergence, we measure the energy distance between the current model distribution $p_u$ the target distribution $p_\text{target}$,
\begin{align*}
    D(p_\text{target}, p_u)^2 = 2 \mathbb{E}_{X,Y}[|| X - Y ||] - \mathbb{E}_{X,X'}[|| X - X'||] - \mathbb{E}_{Y,Y'}[||Y-Y'||],
\end{align*}
where $X, X' \sim p_\text{target}$ and $Y, Y' \sim p_u$ are sampled independently. The energy distance decreases steadily over outer iterations, confirming that fine-tuning drives the learned distribution closer to the target.

\paragraph{Implementation Details}
We use a batch size of $10000$ and no buffer. We perform $10$ fine-tuning steps with $5000$ gradient updates per iteration. We adaptively compute $c$ to ensure a resampling rate of $0.8$. The total training time is approximately $3$ minutes on a single NVIDIA GeForce RTX 4090.

\subsection{Class-conditional sampling on MNIST}
We study class-conditional sampling on MNIST under both smooth and non-smooth reward functions. Let $\vc$ denote a target class and $p(\vc \mid \vx)$ the class probabilities of a pre-trained classifier. In both cases, we aim to sample from a tilted distribution \eqref{eq:tilted_dist}, where the reward pushes samples to match the target class.

\paragraph{Smooth vs.\ non-smooth rewards} Many fine-tuning approaches require the gradient of the reward function. For our importance fine-tuning, we only require evaluations of the reward. We consider two choices of reward.

\emph{Smooth reward.}
For the smooth setting, we define
\[
r(\vx) = \log p(\vc \mid \vx),
\]
which is differentiable with respect to $\vx$ and provides informative gradients. This allows us to exploit the reward-informed architecture in \eqref{eq:lkhd_informed_model}. We compare three methods:
(i) \emph{Online fine-tuning} (Online FT), which minimises the SOC objective \eqref{eq:control_deft} using VarGrad \cite{richter2020vargrad};
(ii) \emph{Classifier guidance}; and
(iii) our proposed \emph{Importance fine-tuning} (Importance FT).

\emph{Non-smooth reward.}
To remove access to reward gradients, we define a binary, non-differentiable reward
\[
r(\vx) =
\begin{cases}
1, & \text{if } \vc = \arg\max_{\vc'} \log p(\vc' \mid \vx), \\
0, & \text{otherwise}.
\end{cases}
\]
This reward is piecewise constant and has zero gradient almost everywhere, rendering gradient-based guidance methods inapplicable. In this case, we parametrise the control as a generic time-dependent network $u_t^\theta(\vx_t)=\mathrm{NN}_1(\vx_t,t)$, without the reward-informed inductive bias. We compare Importance FT against Top-$K$ sampling for different values of $K$. 

\begin{figure}[t]
\centering
\begin{subfigure}[t]{.16\textwidth}
  \includegraphics[width=\linewidth]{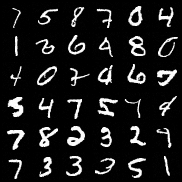}
  \caption*{\footnotesize Prior}
\end{subfigure}\hfill
\begin{subfigure}[t]{.16\textwidth}
  \includegraphics[width=\linewidth]{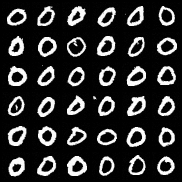}
  \caption*{\footnotesize Zero}
\end{subfigure}\hfill
\begin{subfigure}[t]{.16\textwidth}
  \includegraphics[width=\linewidth]{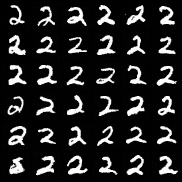}
  \caption*{\footnotesize Two}
\end{subfigure}\hfill
\begin{subfigure}[t]{.16\textwidth}
  \includegraphics[width=\linewidth]{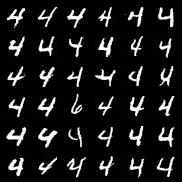}
  \caption*{\footnotesize Four}
\end{subfigure}\hfill
\begin{subfigure}[t]{.16\textwidth}
  \includegraphics[width=\linewidth]{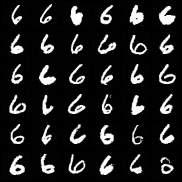}
  \caption*{\footnotesize Six}
\end{subfigure}\hfill
\begin{subfigure}[t]{.16\textwidth}
  \includegraphics[width=\linewidth]{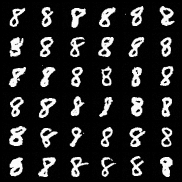}
  \caption*{\footnotesize Eight}
\end{subfigure}
\caption{Class-conditional sampling for MNIST using the smooth reward function: prior, zero, two, four, six, eight.}
\label{fig:mnist_class_cond_1x6}
\end{figure}

Results are reported in Table~\ref{tab:mnist_results_reward_comparison} for the MNIST classes $\{0,2,4,6,8\}$, using expected reward and FID~\cite{heusel2017gans} computed from the penultimate-layer features of the classifier. We show samples in Figure~\ref{fig:mnist_class_cond_1x6}. For smooth rewards, classifier guidance achieves the highest accuracy across all classes, benefiting from access to reward gradients at sampling time. Importance FT consistently improves over Online FT in both accuracy and FID, and in several cases (classes 4 and 6) achieves the best FID among all methods. For the non-smooth reward, importance FT achieves performance close to the smooth-reward setting, with only modest degradation for some classes. Top-$K$ sampling only becomes competitive for very large $K$ (e.g.\ $K=40$), and even then exhibits large variability across classes. Since Top-$K$ requires drawing $K \times N$ samples to obtain $N$ accepted ones, this comparison also highlights the computational inefficiency of rejection-based baselines. Overall, these results show that our method does not rely on differentiable reward functions, while still benefiting from it when available.

\begin{table}[t]
\centering
\caption{Accuracy and FID for class-conditional sampling on MNIST.
Top block: smooth reward functions. Bottom block: non-smooth reward functions.
For Top-$K$, we draw $100$ independent samples.}
\resizebox{\textwidth}{!}{%
\begin{tabular}{lcccccccccc}
\toprule
\textbf{Class} 
& \multicolumn{2}{c}{\textbf{0}} 
& \multicolumn{2}{c}{\textbf{2}} 
& \multicolumn{2}{c}{\textbf{4}} 
& \multicolumn{2}{c}{\textbf{6}} 
& \multicolumn{2}{c}{\textbf{8}} \\
\cmidrule(lr){2-3}\cmidrule(lr){4-5}\cmidrule(lr){6-7}\cmidrule(lr){8-9}\cmidrule(lr){10-11}
& Acc. & FID & Acc. & FID & Acc. & FID & Acc. & FID & Acc. & FID \\
\midrule
\multicolumn{11}{l}{\textbf{Smooth reward functions}} \\
\midrule
Classifier Guid. & \textbf{99.90} & \textbf{16.96} & \textbf{99.90} & \textbf{38.99} & \textbf{100.0} & 25.00 & \textbf{99.90} & 34.54 & \textbf{100.0} & \textbf{8.43} \\
Online FT           & 94.63 & 30.34 & 96.88 & 45.64 & 94.43 & 26.87 & 88.09 & 83.06 & 93.55 & 17.92 \\
Importance FT       & 98.24 & 38.53 & 93.65 & 49.78 & 96.97 & \textbf{16.40} & 91.21 & \textbf{28.32} & 93.06 & 25.36 \\
\midrule
\multicolumn{11}{l}{\textbf{Non-smooth reward functions}} \\
\midrule
Top-$K$ ($k=10$)  & 61.00 & 489.6 & 53.00 & 642.5 & 67.00 & 381.7 & 52.00 & 797.3 & 61.00 & 320.3 \\
Top-$K$ ($k=20$)  & 82.00 & 200.7 & 71.00 & 347.0 & 87.00 & 98.70 & 71.00 & 375.8 & 76.00 & 192.9 \\
Top-$K$ ($k=40$)  & \textbf{96.00} & 43.80 & 88.00 & 161.4 & \textbf{100.0} & \textbf{15.74} & \textbf{95.00} & 62.39 & \textbf{96.00} & 34.93 \\
Importance FT     & 94.43 & \textbf{32.84} & \textbf{92.00} & \textbf{47.47} & 95.61 & 19.69 & 90.72 & \textbf{58.02} & 89.45 & \textbf{28.61} \\
\bottomrule
\end{tabular}}
\label{tab:mnist_results_reward_comparison}
\end{table}

\begin{figure}[t]
\begin{subfigure}[t]{.19\textwidth}
    \includegraphics[width=1.0\linewidth]{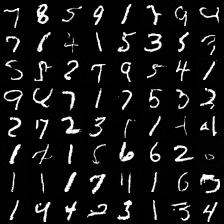}%
    \captionsetup{justification=centering}
    \caption*{$\lambda=100.0$}
\end{subfigure}%
\hfill
\begin{subfigure}[t]{.19\textwidth}
    \includegraphics[width=1.0\linewidth]{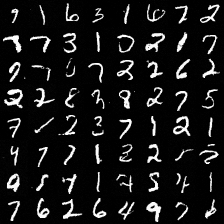}%
        \captionsetup{justification=centering}
    \caption*{$\lambda=10.0$}
\end{subfigure}%
\hfill
\begin{subfigure}[t]{.19\textwidth}
    \includegraphics[width=1.0\linewidth]{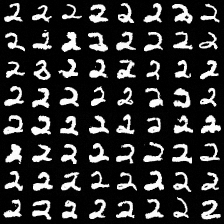}%
            \captionsetup{justification=centering}
    \caption*{$\lambda=1.0$}
\end{subfigure}%
\hfill
\begin{subfigure}[t]{.19\textwidth}
    \includegraphics[width=1.0\linewidth]{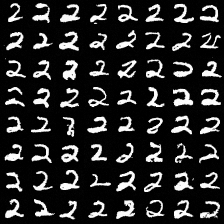}%
            \captionsetup{justification=centering}
    \caption*{$\lambda=0.5$}
\end{subfigure}%
\hfill
\begin{subfigure}[t]{.19\textwidth}
    \includegraphics[width=1.0\linewidth]{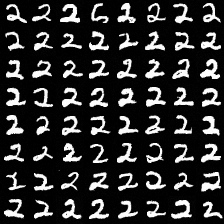}%
            \captionsetup{justification=centering}
    \caption*{$\lambda=0.01$}
\end{subfigure}%
\caption{Samples for changing the temperature $\lambda$ for MNIST class conditional sampling for the digit "Two" using the smooth reward function.}\label{fig:MNIST_change_lambda}
\end{figure}

\paragraph{Diversity--accuracy trade-off}
We next study the effect of the temperature parameter $\lambda$ in the tilted distribution \eqref{eq:tilted_dist}, which controls the trade-off between fidelity to the base model and maximising the reward. 
Smaller values of $\lambda$ place more weight on the reward, increasing accuracy but reducing diversity. Quantitative results are shown in Table~\ref{tab:results_mnist_lambda} for class ``Two'', reporting FID, expected reward, and classification accuracy on $1024$ samples; visual samples are shown in Figure~\ref{fig:MNIST_change_lambda}. For large $\lambda$ (e.g.\ $\lambda=100.0$), the tilted distribution is close to the base model and the classifier has little effect, resulting in low accuracy. The best FID is achieved at an intermediate temperature ($\lambda=0.5$). For small $\lambda$ (e.g.\ $\lambda=0.01$), accuracy increases but diversity collapses, leading to worse FID. Thus, we can control the trade-off between diversity and accuracy using the temperature $\lambda$. As discussed in Section \ref{sec:2Dtoyexample}, this is not possible to reward-only fine-tuning methods, which rank trajectory only based on the raw reward value and not consider closeness to the base model in the resampling criterion.

\begin{table}[t]
\centering
\caption{Results for varying $\lambda$ for class conditional sampling on MNIST for class $2$.}
\begin{tabular}{lccccccc}
\toprule $\lambda$ & $100.0$ & $10.0$ & $1.0$ & $0.5$ & $0.25$ & $0.1$ & $0.01$ \\ \midrule
 FID $(\downarrow)$ &  $1457.99$ & $934.26$ & $123.51$ & $84.96$ & $122.25$ & $151.28$ & $149.72$  \\
  $\mathbb{E}[r(x)]$ $(\uparrow)$ & $-13.54$  & $-8.49$ & $-0.094$ & $-0.078$ & $-0.102$ & $-0.043$ & $-0.062$ \\
 Accuracy $(\uparrow)$& $7.91$ & $17.28$ & $98.53$ & $98.34$ & $98.53$  & $99.12$ & $99.02$  \\ \bottomrule  
\end{tabular}
\label{tab:results_mnist_lambda}
\end{table}

\paragraph{Implementation details}
We pre-train a score-based diffusion model on MNIST using a VP-SDE with $\beta_{\min}=0.1$ and $\beta_{\max}=20.0$. The score network is a small Attention UNet \cite{dhariwal2021diffusion} with approximately $1.1$M parameters, and the control network has approximately $0.8$M parameters. Importance FT is run for $50$ iterations with $50$ gradient updates per iteration, batch size $256$, buffer size $2048$, reward scaling $\lambda=4.0$, and KL regularisation weight $\alpha_{\mathrm{KL}}=0.001$. The acceptance threshold is chosen adaptively, accepting $10\%$ of samples for the first $10$ iterations and $30\%$ thereafter. Total training time is approximately $10$ minutes on a single NVIDIA GeForce RTX~4090. For classifier guidance we use a guidance scale $\gamma=4.5$, and for Online FT we use VarGrad with the formulation in \cite[Appendix G.3]{denker2024deft}. All methods share the same underlying diffusion model and evaluation script.

\subsection{Posterior Sampling for Inverse Problems}
\begin{figure*}[t]
\centering
\includegraphics[width=0.9\linewidth]{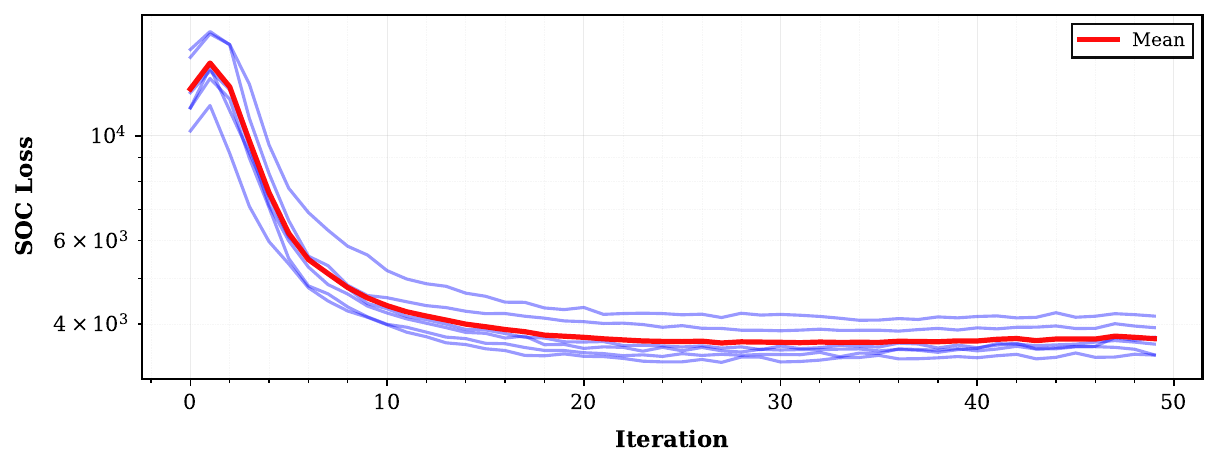}
\caption{Posterior sampling on \texttt{Flowers}. The SC loss $\mathcal{L}_\text{SC}(h_k)$ decreases over fine-tuning iterations $k$. Red shows the mean; blue shows individual runs for observations $y^{(i)}$.} \label{fig:superresolution}
\end{figure*}

We consider posterior sampling in inverse, where the reward is defined by the likelihood $r(\vx; \vy) = \log p^\text{lkhd}(\vy | \vx)$, i.e., the conditional distributions of measurements $\vy$ given images $\vx$. In particular, we focus on super-resolution, where observations are generated as $\vy =\fA \vx + \eta$, with $\eta \sim \mathcal{N}(0, \sigma_y^2 \mathbf{I})$ with $\fA$ denoting the downsampling operator. The resulting log-likelihood is 
\begin{align}
    r(\vx; \vy) = -\frac{1}{2\sigma_y^2}\|\fA \vx - \vy \|_2^2.
\end{align}
At the start of fine-tuning, the terminal states $\vx_0$ of most sampled trajectories are inconsistent with the measurements $\vy$, yielding low reward. To stabilise fine-tuning, we amortise over measurements by sampling synthetic observations $\tilde \vy \sim p^{\text{lkhd}}(\tilde \vy \mid \vx_0)$ and conditioning the control on $\tilde \vy$. This leads to the amortised fine-tuning objective
\begin{align}
\mathcal L_{\mathrm{FT}}(u) = \mathbb{E}_{\substack{
\vx_0 \sim \tilde{\sP}_{h^k}^0,\;\tilde \vy \sim p^{\text{lkhd}}(\tilde \vy|\vx_0) \\
t \sim \mathrm{U}(0,T),\; \vx_t \sim \fwd{p}_{t|0}(\cdot|\vx_0)
}}
\!\left[\left\|\left(u_t(\vx_t,\tilde \vy)+s_t(\vx_t)\right)-\nabla_{\vx_t}\ln \fwd{p}_{t|0}(\vx_t|\vx_0)\right\|^2\right].
\end{align}
We train a diffusion model on the \texttt{Flowers} dataset~\cite{nilsback2008automated} for a $2\times$ super-resolution task with $5\%$ relative Gaussian noise, using the likelihood-informed architecture~\eqref{eq:lkhd_informed_model}, see Appendix \ref{app:posterior} for details. Theorem~\ref{the:} shows that iterative importance fine-tuning acts as a descent algorithm for the stochastic control (SC) loss $\mathcal L_{\mathrm{SC}}$. Figure~\ref{fig:superresolution} provides numerical verification of this result, demonstrating that the SC loss (free energy) decreases monotonically over fine-tuning iterations.

\begin{figure*}[t]
\centering
\begin{subfigure}[t]{.19\textwidth}
  \includegraphics[width=\linewidth]{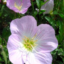}
    \captionsetup{justification=centering}
  \caption*{Ground truth}
\end{subfigure}%
\hfill
\begin{subfigure}[t]{.19\textwidth}
  \includegraphics[width=\linewidth]{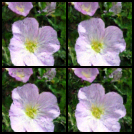}
    \captionsetup{justification=centering}
  \caption*{Importance FT \\ $\mathbb{E}[r(x)]=-15.28$ }
\end{subfigure}%
\hfill
\begin{subfigure}[t]{.19\textwidth}
  \includegraphics[width=\linewidth]{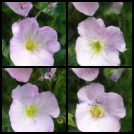}
    \captionsetup{justification=centering}
  \caption*{DPS \\ $\mathbb{E}[r(x)]=-78.95$ }
\end{subfigure}%
\hfill
\begin{subfigure}[t]{.19\textwidth}
  \includegraphics[width=\linewidth]{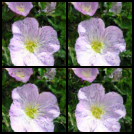}
    \captionsetup{justification=centering}
  \caption*{VarGrad \\ $\mathbb{E}[r(x)]=-18.28$}
\end{subfigure}%
\hfill
\begin{subfigure}[t]{.19\textwidth}
  \includegraphics[width=\linewidth]{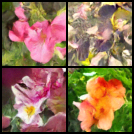}
    \captionsetup{justification=centering}
  \caption*{Top-K \\ $\mathbb{E}[r(x)]=-1619.74$  }
\end{subfigure}%
\caption{Comparison of samples for the super-resolution task on the \texttt{Flowers} dataset.} \label{fig:comparison_flowers}
\end{figure*}

In Figure~\ref{fig:comparison_flowers}, we compare importance fine-tuning against DPS~\cite{chung2022diffusion}, Top-$K$ sampling, and an optimal-control baseline using VarGrad~\cite{richter2020vargrad}. Here, importance fine-tuning achieves the highest reward. DPS produces overly smooth samples, while Top-$K$ sampling fails due to the narrow reward landscape. VarGrad yields visually similar samples but attains a lower reward on average. In Figure \ref{fig:flower_additional_results} (Appendix \ref{app:posterior}) we additionally present the per-pixel mean and standard deviation.

\subsection{Text-to-Image Reward Fine-tuning}
Despite the progress in training text-to-image diffusion models, samples do not always align with human preferences. In particular, the model can struggle with unnatural prompts such as ``A green colored rabbit'' \cite{venkatraman2024amortizing,fan2024reinforcement}. To address this, a common strategy is to fine-tune the model to maximise a reward signal that approximates human judgment. We evaluate our approach on Stable Diffusion-v1.5 \cite{rombach2022high}, aligning it via the ImageReward-v1.0 \cite{xu2024imagereward} preference model, see Appendix~\ref{app:rewardfinetuning} for experimental details. 

\paragraph{Baselines and Comparisons} We compare our method against two inference-time strategies: Top-$K$ sampling (with $K=6$), which simply selects the highest-reward candidate from a batch, and FK-Steering \cite{singhal2025a}. Furthermore, we benchmark against state-of-the-art fine-tuning approaches: DPOK \cite{fan2024reinforcement}, Relative Trajectory Balance (RTB) \cite{venkatraman2024amortizing}, and Adjoint Matching \cite{domingo2024adjoint}. We adopt the LoRA parameterisation \cite{hu2021lora} for DPOK, RTB, and our method. For Adjoint Matching, we report results for both the standard full-model fine-tuning and a LoRA variant.

\paragraph{Results}
Quantitative results are summarised in Table~\ref{tab:results_stablediff}, with samples provided in Figure~\ref{fig:stable_diff_greenrabbit}, see Appendix~\ref{app:rewardfinetuning} for additional samples. We report both the mean reward and diversity, which is measured as $1 - \text{cosine similarity}$ of CLIP embeddings \cite{radford2021learning}. Across three prompts, all fine-tuning methods improve rewards relative to the base model. Adjoint Matching achieves the highest overall reward (e.g., $1.706$ for ``A green colored rabbit''). However, this comes at the expense of diversity and a significantly higher memory requirement due to full-model adaptation. In contrast, Importance Fine-Tuning achieves competitive rewards while maintaining diversity levels comparable to DPOK and RTB. Crucially, our method shares the efficient $O(B)$ memory footprint of DPOK and, like RTB, requires only reward evaluations without gradient access. Overall, importance fine-tuning offers more of a middle ground; achieving a lower reward than adjoint matching but with a more efficient and scalable training pipeline, making it an option for researchers without access to large-scale compute.

\begin{figure}[t]
\begin{subfigure}[t]{.19\textwidth}
    \includegraphics[width=0.5\linewidth]{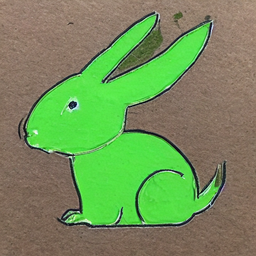}%
    \includegraphics[width=0.5\linewidth]{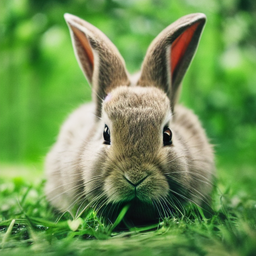}
\end{subfigure}%
\hfill
\begin{subfigure}[t]{.19\textwidth}
    \includegraphics[width=0.5\linewidth]{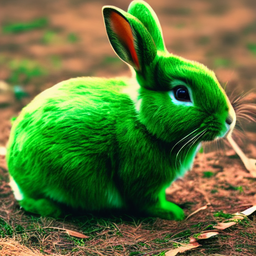}%
    \includegraphics[width=0.5\linewidth]{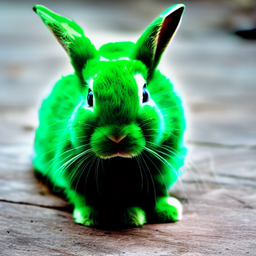}
\end{subfigure}%
\hfill
\begin{subfigure}[t]{.19\textwidth}
    \includegraphics[width=0.5\linewidth]{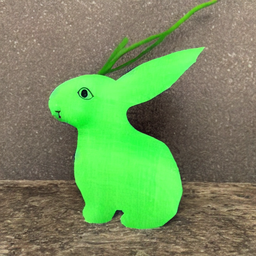}%
    \includegraphics[width=0.5\linewidth]{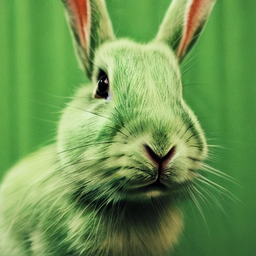}
\end{subfigure}%
\hfill
\begin{subfigure}[t]{.19\textwidth}
    \includegraphics[width=0.5\linewidth]{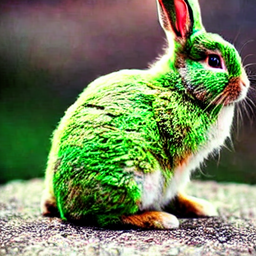}%
        \includegraphics[width=0.5\linewidth]{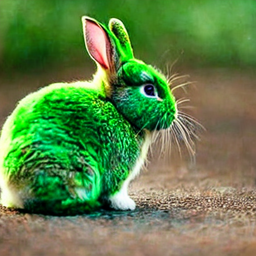}
\end{subfigure}%
\hfill
\begin{subfigure}[t]{.19\textwidth}
    \includegraphics[width=0.5\linewidth]{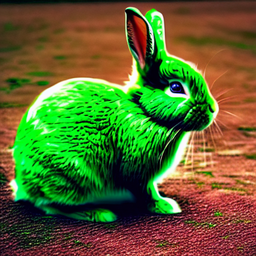}%
        \includegraphics[width=0.5\linewidth]{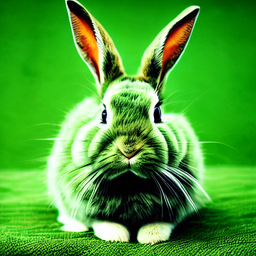}
\end{subfigure}%

\begin{subfigure}[t]{.19\textwidth}
    \includegraphics[width=0.5\linewidth]{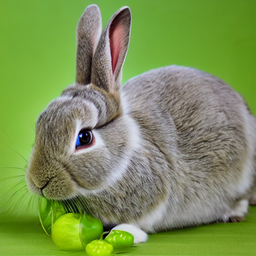}%
    \includegraphics[width=0.5\linewidth]{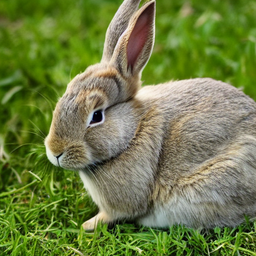}
     \captionsetup{justification=centering}
    \caption*{Base Model}
\end{subfigure}%
\hfill
\begin{subfigure}[t]{.19\textwidth}
    \includegraphics[width=0.5\linewidth]{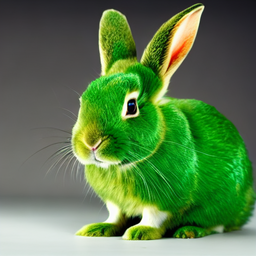}%
    \includegraphics[width=0.5\linewidth]{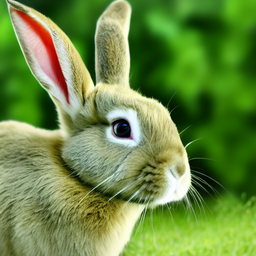}
    \captionsetup{justification=centering}
  \caption*{DPOK}
\end{subfigure}%
\hfill
\begin{subfigure}[t]{.19\textwidth}
    \includegraphics[width=0.5\linewidth]{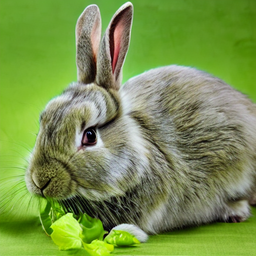}%
    \includegraphics[width=0.5\linewidth]{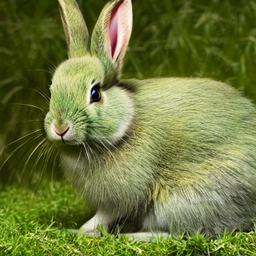}
    \captionsetup{justification=centering}
  \caption*{RTB}
\end{subfigure}%
\hfill
\begin{subfigure}[t]{.19\textwidth}
    \includegraphics[width=0.5\linewidth]{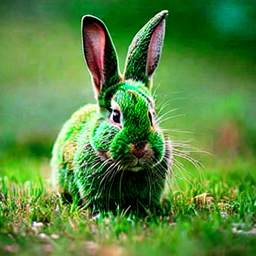}%
    \includegraphics[width=0.5\linewidth]{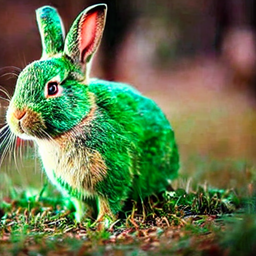}
    \captionsetup{justification=centering}
  \caption*{Adjoint Matching}
\end{subfigure}%
\hfill
\begin{subfigure}[t]{.19\textwidth}
    \includegraphics[width=0.5\linewidth]{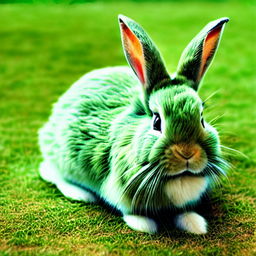}%
    \includegraphics[width=0.5\linewidth]{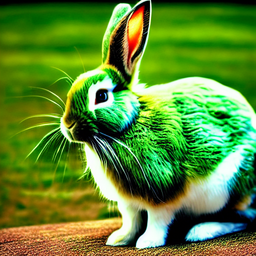}
    \captionsetup{justification=centering}
  \caption*{Importance FT}
\end{subfigure}%
\caption{Samples for the base model, DPOK, Adjoint Matching and our importance FT for the prompt ''A green colored rabbit.``. Images were generated using the same seed.} \label{fig:stable_diff_greenrabbit}
\end{figure}

\begin{table}[h]
\centering
\caption{Results for text-to-image. We compute the mean reward and diversity over $100$ samples. For the GPU memory $B$ refers to the batch size and $T$ to the number of time steps during sampling.}
\resizebox{\textwidth}{!}{%
\begin{tabular}{lcccccccc}
\toprule                       & \multicolumn{2}{c}{\textbf{``A green colored rabbit.``}} & \multicolumn{2}{c}{\textbf{``Two roses in a vase.``}} & \multicolumn{2}{c}{\textbf{``Two dogs in the park.``}} & $\nabla r(x)$ free & GPU memory  \\ 
                 & Reward $(\uparrow)$  & Diversity $(\uparrow)$ & Reward $(\uparrow)$ & Diversity $(\uparrow)$ & Reward $(\uparrow)$ & Diversity $(\uparrow)$ & & (training) \\ \midrule
{\small \textbf{Base Model}} & $-0.207$ & $0.154$ & $1.130$ & $0.108$ & $0.518$ &  $0.193$ &  \cmark & \text{N/A} \\ 
{\small \textbf{Top-K Sampling} ($K=6$)}  & $1.334$ & $0.149$  & $1.619$ & $0.121$ & $0.970$ & $0.161$ & \cmark & \text{N/A} \\ 
{\small \textbf{FK-Steering} ($K=6$)}  & $1.207$ & $0.147$  & $1.645$ & $0.108$ & $0.985$ & $0.139$ & \cmark & \text{N/A} \\ 
{\small \textbf{DPOK}}  & $1.621$ & $0.065$ & $1.592$ & $0.106$ & $1.013$ & $0.144$ & \cmark   & $O(B)$  \\  
{\small \textbf{RTB}} & $1.435$ & $0.092$  & $1.501$ & $0.127$  & $1.058$ & $0.174$ & \cmark & $O(BT)$ \\  
{\small \textbf{Adjoint Matching}} & $1.706$ & $0.050$ & $1.504$ & $0.088$ & $1.325$ & $0.132$ & \xmark &  $O(B)$ \\ 
{\small \textbf{Adjoint Matching} (LoRA)} & $1.435$ & $0.092$ & $1.230$ & $0.137$ & $0.780$ & $0.162$ & \xmark &  $O(B)$ \\ 
{\small \textbf{Importance FT (ours)}} & $1.462$ & $0.051$ & $1.534$ & $0.077$ & $1.010$ & $0.120$ & \cmark &  $O(B)$ \\ \bottomrule
\end{tabular}}
\label{tab:results_stablediff}
\end{table}

\begin{figure}[t]
\begin{subfigure}[t]{.24\textwidth}
    \includegraphics[width=0.5\linewidth]{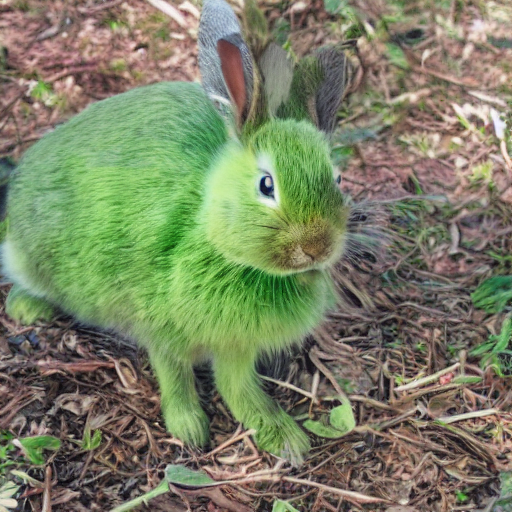}%
    \includegraphics[width=0.5\linewidth]{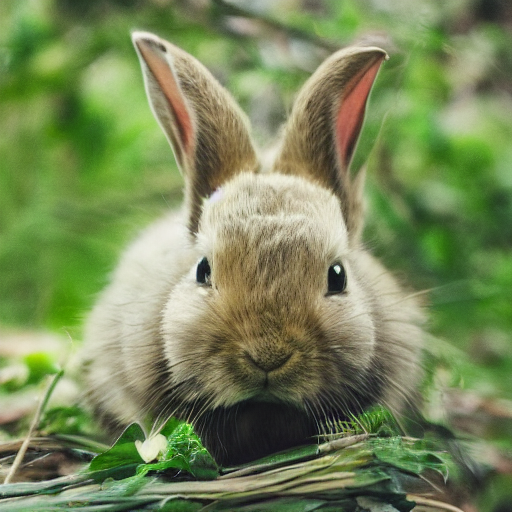}
\end{subfigure}%
\hfill
\begin{subfigure}[t]{.24\textwidth}
    \includegraphics[width=0.5\linewidth]{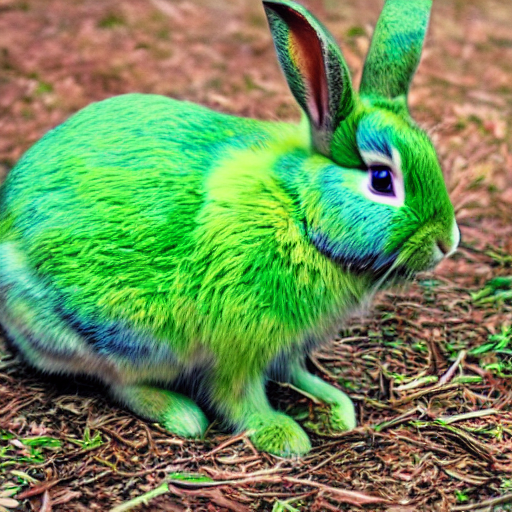}%
    \includegraphics[width=0.5\linewidth]{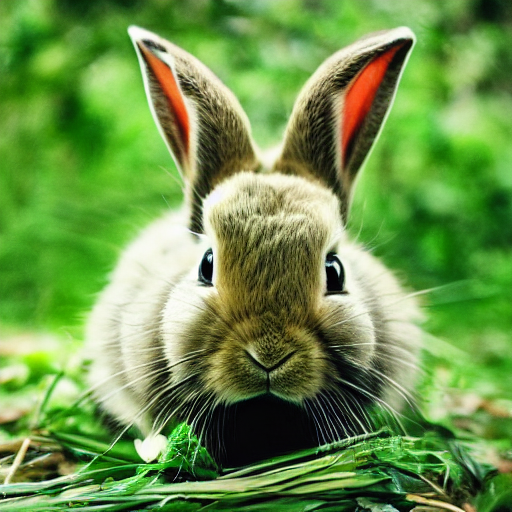}
\end{subfigure}%
\hfill
\begin{subfigure}[t]{.24\textwidth}
    \includegraphics[width=0.5\linewidth]{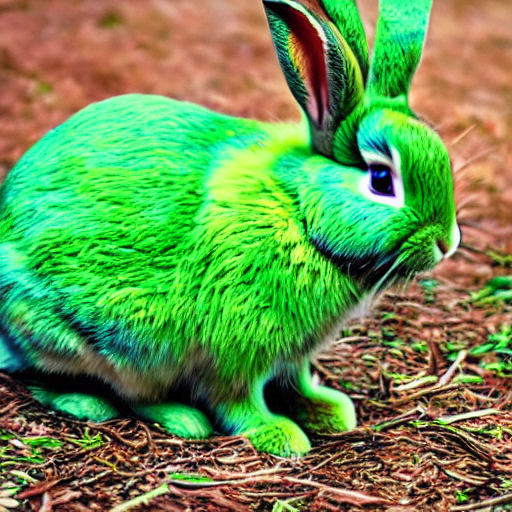}%
        \includegraphics[width=0.5\linewidth]{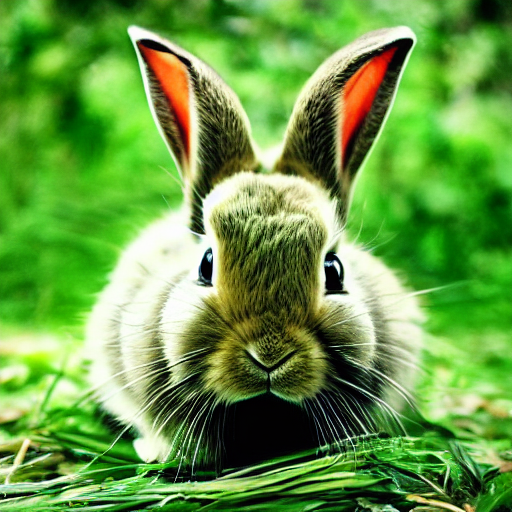}
\end{subfigure}%
\hfill
\begin{subfigure}[t]{.24\textwidth}
    \includegraphics[width=0.5\linewidth]{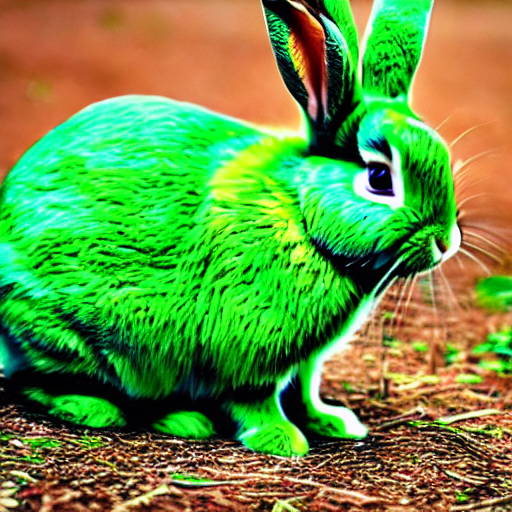}%
        \includegraphics[width=0.5\linewidth]{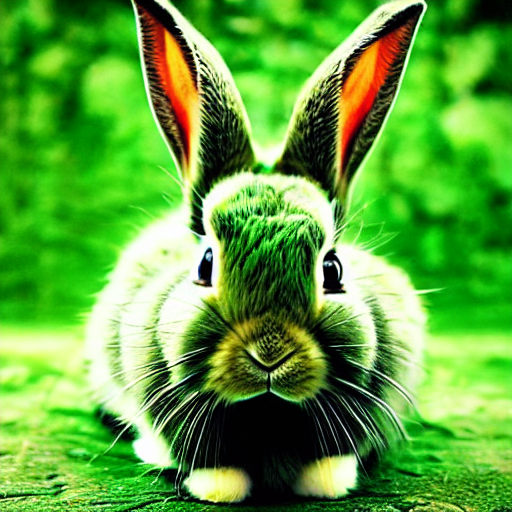}
\end{subfigure}%

\begin{subfigure}[t]{.24\textwidth}
    \includegraphics[width=0.5\linewidth]{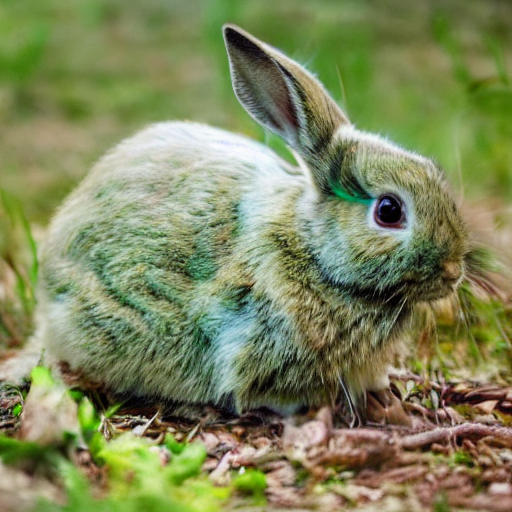}%
    \includegraphics[width=0.5\linewidth]{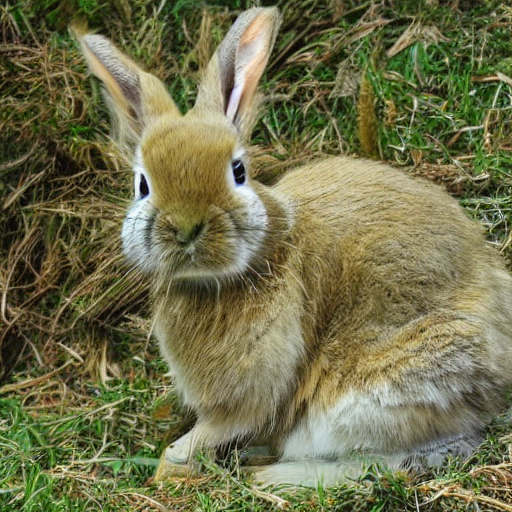}
     \captionsetup{justification=centering}
    \caption*{$w=2.0$}
\end{subfigure}%
\hfill
\begin{subfigure}[t]{.24\textwidth}
    \includegraphics[width=0.5\linewidth]{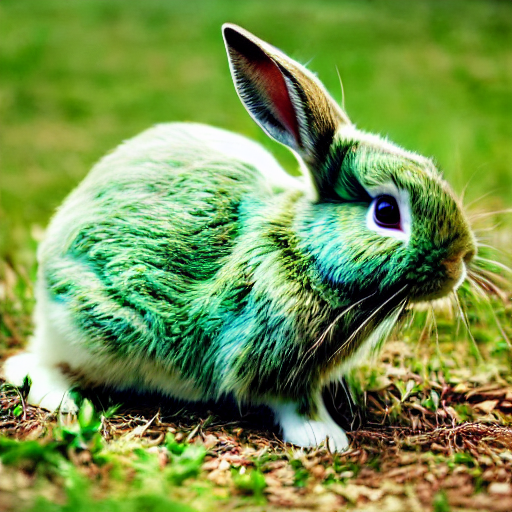}%
    \includegraphics[width=0.5\linewidth]{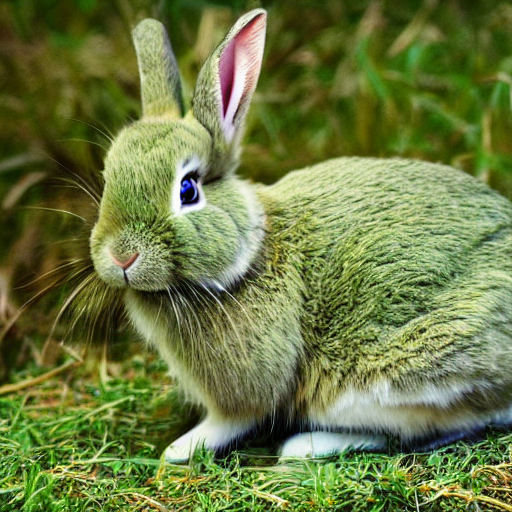}
    \captionsetup{justification=centering}
  \caption*{$w=4.0$}
\end{subfigure}%
\hfill
\begin{subfigure}[t]{.24\textwidth}
    \includegraphics[width=0.5\linewidth]{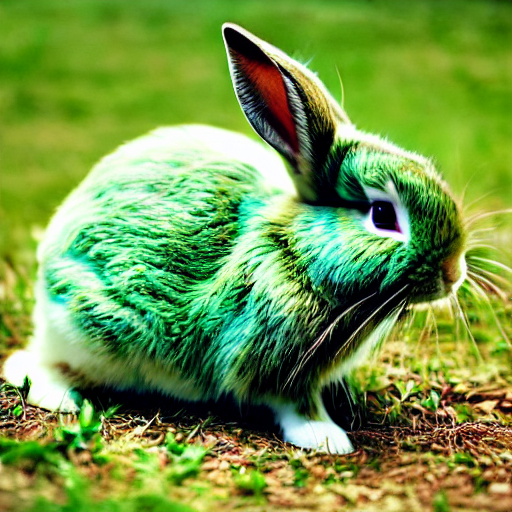}%
        \includegraphics[width=0.5\linewidth]{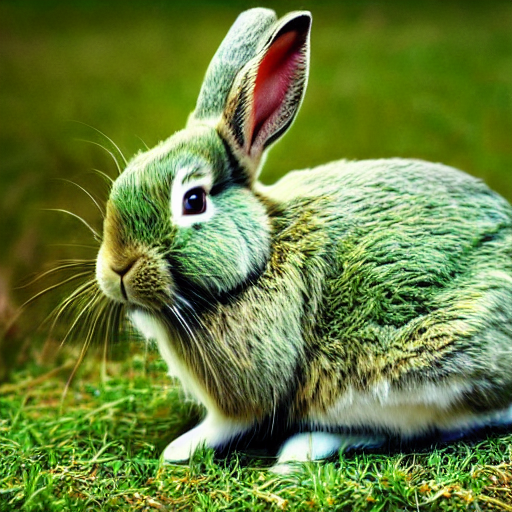}
    \captionsetup{justification=centering}
  \caption*{$w=5.0$}
\end{subfigure}%
\hfill
\begin{subfigure}[t]{.24\textwidth}
    \includegraphics[width=0.5\linewidth]{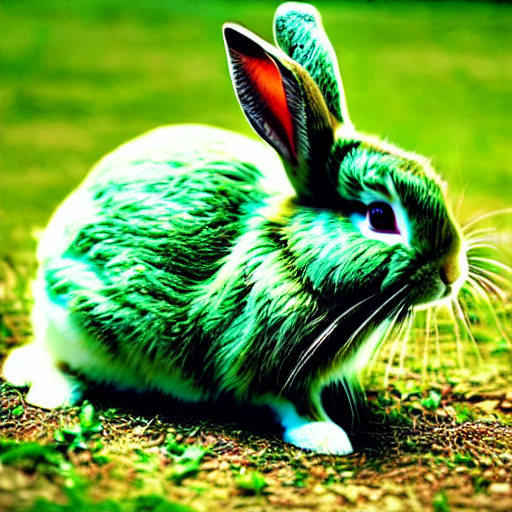}%
        \includegraphics[width=0.5\linewidth]{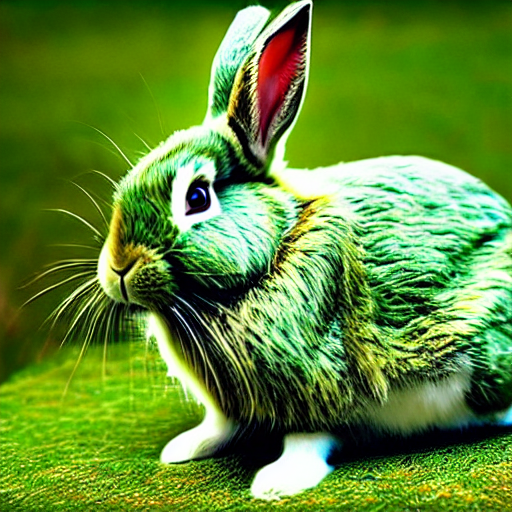}
    \captionsetup{justification=centering}
  \caption*{$w=7.5$}
\end{subfigure}%
\caption{Samples for importance FT for the prompt ''A green colored rabbit.`` with different guidance scales $w$ for classifier free guidance. Images were generated using the same seed.} \label{fig:stable_diff_greenrabbit_guidance_scale}
\end{figure}

\paragraph{Guidance Scale Sensitivity}
Text-to-image diffusion models typically employ classifier-free guidance (CFG). In CFG both an unconditional $\epsilon^\theta_t(\vx_t)$ and a conditional $\epsilon^\theta_t(\vx_t, c)$ noise-prediction model are learned by randomly masking out the text prompt $c$. During sampling the linear combination $\bar{\epsilon}^\theta_t(\vx_t, t) = (1 + w) \epsilon^\theta_t(\vx_t, c) - w \epsilon^\theta_t(\vx_t)$, with a guidance scale $w$, is used. A critical distinction in our approach is the handling of CFG guidance scale $w$. While RL-based methods, such as Adjoint Matching or DPOK, typically fix the scale during training, our score-matching loss is independent of the guidance scale, allowing it to be tuned at inference time. We observe that the default Stable Diffusion guidance scale ($w=7.5$) can produce oversaturated outputs after fine-tuning. As shown in Figure~\ref{fig:stable_diff_greenrabbit_guidance_scale}, tuning the guidance scale post-training yields cleaner, more natural images.

\section{Conclusion}\label{sec:concl}
We propose an iterative approach for fine-tuning diffusion models for conditional sampling tasks. As the naive iterative retraining of generative models often lead to performance degradation \cite{shumailov2023curse}, we introduce an additional resampling step based on path-based importance weights. We present experiments for class conditional sampling, inverse problems and reward fine-tuning of text-to-image diffusion models. Online fine-tuning methods often require additional tricks to increase training stability, e.g., \cite{venkatraman2024amortizing} use loss clipping and disregard low reward trajectories or \cite{fan2024reinforcement} employ variance reduction techniques by additionally learning the value function. In contrast our importance fine-tuning method is trained using a score matching loss, see Theorem~\ref{th:deft}, leading to a more stable training.  

\paragraph{Limitations} In our approach, the diffusion model is fine-tuned using ``good'' samples, i.e., having a high importance weight. However, these samples necessarily lie in the support of the pre-trained model and we rely on the fact that such ``good'' samples exist. Thus, our refinement approach might struggle on domains where the distribution of the pre-trained model is sparse and high reward samples are. One possibility to alleviate this problem is to make use of \textit{off-policy} samples \cite{venkatraman2024amortizing}, which are obtained by some other method. For off-policy samples, we loose the compact formulation of the RND from Lemma~\ref{lemma:rnd}. However, we can still compute importance weights using the RND in Proposition~\ref{prop:rnd_between_sde}, see Appendix \ref{app:RND_SDE}.

\appendix
\section{Radon-Nikodym derivative betweens SDEs}
\label{app:RND_SDE}
To calculate the approximated importance weights, we make use of the RND between SDEs. In particular, we use the result from \cite{nusken2021solving,nusken2024transport}.

\begin{proposition}(RND between SDEs \cite{nusken2021solving,nusken2024transport})
\label{prop:rnd_between_sde}
Given the following SDEs 
\begin{align} 
    \dd \bY_t &=  a_t(\bY_t) \,\dd t + \sigma_t(\bY_t)\; \fwd{ \dd \rv{W}}_t,  \quad \bY_0 \ \sim  \mu , \\
    \dd \bX_t &=  b_t(\bX_t) \,\dd t + \sigma_t(\bX_t) \; \fwd{ \dd \rv{W}}_t,  \quad \bX_0 \ \sim  \nu ,
\end{align}
with path probabilities $\fwd{\sP}^a$ and $\fwd{\sP}^b$, we get 
\begin{align} 
    \ln \left( \frac{d \fwd{\sP}^a}{d\fwd{\sP}^b}\right)(\bZ) = \ln \left( \frac{d \mu}{d \nu} \right)(\bZ_0) + \int_0^T \sigma_t^{-2} (a_t - b_t)(\bZ_t) d \fwd{\bZ}_t + \frac12 \int_0^T \sigma_t^{-2} (b_t^2 - a_t^2)(\bZ_t) dt,
\end{align}
and in particular, when evaluated on $\bY$:
\begin{align} 
    \ln \left( \frac{d \fwd{\sP}^a}{d\fwd{\sP}^b}\right)(\bY) = \ln \left( \frac{d \mu}{d \nu} \right)(\bY_0) + \int_0^T \sigma_t^{-1} (a_t - b_t)(\bY_t) d \fwd{\bW}_t + \frac12 \int_0^T \sigma_t^{-2} \|b_t - a_t\|^2(\bY_t) dt.
\end{align}
\end{proposition}

Let $\sP^\text{data}$ be the path probability of 
\begin{align}
    \dd \bX_t &= \left( f_t(\bX_t) - \sigma_t^2 \nabla_{\bX_t} \ln p_t(\bX_t)  \right) \,\dd t + \sigma_t \bwd{ \dd \rv{W}}_t, \quad \bX_T \sim  \sP_T, 
\end{align}
and $\sP^u$ the path probability of 
\begin{align}
    \dd \bH_t &= \left( f_t(\bH_t) - \sigma_t^2( \nabla_{\bH_t} \ln p_t(\bH_t) + u_t(\bH_t)) \right) \,\dd t + \sigma_t \bwd{ \dd \rv{W}}_t, \quad \bH_T \sim  \sP_T. 
\end{align}
For the approximated importance weights we require the RND of $\sP^u$ with respect to $\sP^\text{data}$ evaluated at trajectories from $\sP^u$. Let the drift of $\sP^\text{data}$ be given as $a_t(\vx) = f_t(\vx) - \sigma_t^2 \nabla_{\vx} \ln p_t(\vx)$ and the drift of $\sP_u$ as $b_t(\vx) = f_t(\vx) - \sigma_t^2 \nabla_{\vx} \ln p_t(\vx) - \sigma_t^2 u_t(\vx) $. In particular, we have $a_t - b_t = \sigma_t^2 u_t$. For readability, we do not omit the dependence on $\vx$ for the drift in the following. Using Proposition \ref{prop:rnd_between_sde} we get,
\begin{align}
    \ln\left( \frac{\dd \sP^\text{data}}{\dd \sP^u} \right)(\bH) &= \int_0^T \sigma_t^{-2} (a_t - b_t) \dd \bwd{\bH}_t + \frac12 \int_0^T \sigma_t^{-2} (b_t^2 - a_t^2) \dd t \\ 
    &= \int_0^T \sigma_t^{-2}(a_t-b_t)b_t \dd t + \int_0^T \sigma_t^{-2} (a_t - b_t) \sigma_t d \bwd{\bW}_t + \frac12 \int_0^T \sigma_t^{-2} (b_t^2 - a_t^2) \dd t \\
    &= \frac12 \int_0^T \sigma_t^{-2} [2a_t b_t - 2b_t^2 +  b_t^2 -  a_t^2] \dd t + \int_0^T \sigma_t u_t \dd \bwd{\bW}_t \\ 
    & = - \frac12 \int_0^T \sigma_t^2 \| u_t \|_2^2 \dd t + \int_0^T \sigma_t u_t \dd \bwd{\bW}_t, \label{eq:rnd_final}
\end{align}
evaluated on a trajectory $\bH$ from $\sP^u$.

\subsection{Discrete Version for DDPM}
We can also express the RND in the discrete setting. For this derivation, we make use of the DDPM schedule \cite{ho2020denoising}. The forward diffusion process in DDPM is defined by a variance schedule $\{\beta_t \}_{t=1}^T$ that gradually transforms the data into a Gaussian distribution through transitions $q(\mathbf{x}_t | \mathbf{x}_{t-1}) = \mathcal{N}(\mathbf{x}_t; \sqrt{1-\beta_t}\mathbf{x}_{t-1}, \beta_t \mathbf{I})$. We further define $\alpha_t := 1 - \beta_t$ and the cumulative product $\bar{\alpha}_t := \prod_{s=1}^t \alpha_s$. The generalisation to different schedules is straightforward. The path probability of the pre-trained model is given as 
\begin{align}
    p_\theta^\text{data}(\vx_0, \dots, \vx_T) = p_T(\vx_T) \prod_{t=1}^T p_\theta^\text{data}(\vx_{t-1} | \vx_t), \quad p_\theta^\text{data}(\vx_{t-1}|\vx_t) = \mathcal{N}(\vx_{t-1}| \mu_\theta(\vx_t, t) ; \tilde \beta_t^2 I)
\end{align}
where the mean is parametrised as $\mu_\theta(\vx_t, t) =\frac{1}{\sqrt \alpha_t} ( \vx_t - \frac{1 - \alpha_t}{\sqrt{1 - \bar \alpha_t}} \epsilon_\theta(\vx_t,t))$. Similar, we have path probabilities for the fine-tuned model, using $\epsilon_\theta(\vx_t,t)+u_\varphi(\vx_t,t)$ as the denoising model, as 
\begin{align}
        p_\varphi^u(\vx_0, \dots, \vx_T) = p_T(\vx_T) \prod_{t=1}^T p_\varphi^u(\vx_{t-1} | \vx_t), \quad p_\varphi^u(\vx_{t-1}|\vx_t) = \mathcal{N}(\vx_{t-1}| \mu_u(\vx_t, t) ; \tilde \beta_t^2 I),
\end{align}
with mean $\mu_u(\vx_t, t) = \mu_\theta(\vx_t, t) + \Delta_u(\vx_t, t)$ is the original mean plus a delta given by the control. We use $\tilde \beta_t = \sqrt{\frac{1 - \bar \alpha_{t-1}}{1 - \bar \alpha_t} \beta_t}$ for the standard deviation of the reverse kernel. Using this setting, we can write the RND as
\begin{align}
    \ln \left( \frac{p_\theta^\text{data}(\vx_0, \dots, \vx_T)}{p_\varphi^u(\vx_0, \dots, \vx_T)} \right) = \sum_{t=1}^T  \log \left( \frac{p_\theta^\text{data}(\vx_{t-1}|\vx_t)}{p_\varphi^u(\vx_{t-1}|\vx_t)} \right),
\end{align}
where we assumed that the terminal distribution $p_T$ is the same for both diffusion models. The log ratio of two Gaussian reduces to 
\begin{align}
    \log \left( \frac{\mathcal{N}(x; \mu_1, \Sigma_1)}{\mathcal{N}(x; \mu_2, \Sigma_2)} \right) = - \frac{1}{2} [(x - \mu_1)^T \Sigma_1^{-1}(x - \mu_1) - (x - \mu_2)^T \Sigma_2^{-1}(x - \mu_2)] + \frac12 \log \frac{|\Sigma_2|}{|\Sigma_1|},
\end{align}
which gets us
\begin{align}
    \label{eq:log_ratio_kernel}
    \log \left( \frac{p_\theta^\text{data}(\vx_{t-1}|\vx_t)}{p_\varphi^u(\vx_{t-1}|\vx_t)} \right) = - \frac{1}{2 \tilde \beta_t^2} \left[ \| \vx_{t-1} - \mu_\theta(\vx_t,t) \|_2^2 - \| \vx_{t-1} - \mu_u(\vx_t,t) \|_2^2 \right].
\end{align}
To further simplify these terms, we need some information about the trajectory $\vx_0, \dots, \vx_T$. In particular, we assume that we have a \textit{trajectory sampled from the fine-tuned model}, i.e.,
\begin{align}
    \vx_{t-1} = \mu_u(\vx_t, t) + \tilde \beta_t \epsilon, \quad \epsilon \sim \mathcal{N}(0,I).
\end{align}
Given the parametrisation of the mean in the diffusion model
\begin{align}
    \mu_\theta(\vx_t,t) &=\frac{1}{\sqrt \alpha_t} \left( \vx_t - \frac{1 - \alpha_t}{\sqrt{1 - \bar \alpha_t}} \epsilon_\theta(\vx_t,t)\right), \\
    \mu_u(\vx_t,t) &=\frac{1}{\sqrt \alpha_t} \left( \vx_t - \frac{1 - \alpha_t}{\sqrt{1 - \bar \alpha_t}} \epsilon_\theta(\vx_t,t) - \frac{1 - \alpha_t}{\sqrt{1 - \bar \alpha_t}} u_\varphi(\vx_t,t)\right), \label{eq:ddpm_mean_h}
\end{align}
we can reduce the terms on the RHS of Equation \eqref{eq:log_ratio_kernel} to
\begin{align}
    & \| \vx_{t-1} - \mu_\theta(\vx_t,t) \|_2^2 -  \| \vx_{t-1} - \mu_u(\vx_t,t) \|_2^2 \\ &= \| \mu_u(\vx_t, t) + \tilde \beta_t \epsilon - \mu_\theta(\vx_t, t) \|_2^2 - \| \mu_u(\vx_t, t) + \tilde \beta_t \epsilon - \mu_u(\vx_t, t) \|_2^2 \\
    & = \|  \Delta_u(\vx_t, t) + \tilde \beta_t \epsilon \|_2^2 - \| \tilde \beta_t  \epsilon \|_2^2. \label{eq:mean_difference}
\end{align}
Combining Equation~\eqref{eq:mean_difference} and Equation~\eqref{eq:log_ratio_kernel}, we obtain
\begin{align}
     \log \left( \frac{p_\theta^\text{data}(\vx_{t-1}|\vx_t)}{p_\varphi^u(\vx_{t-1}|\vx_t)} \right) &= - \frac{1}{2 \tilde \beta_t^2} \left[ \| \vx_{t-1} - \mu_\theta(\vx_t,t) \|_2^2 - \| \vx_{t-1} - \mu_u(\vx_t,t) \|_2^2 \right] \\
    & = - \frac{1}{2 \tilde \beta_t^2} \left[\|  \Delta_u(\vx_t, t) + \tilde \beta_t \epsilon \|_2^2 - \| \tilde \beta_t \epsilon \|_2^2\right] \\
    & = - \frac{1}{2 \tilde \beta_t^2} \left[ \| \Delta_u(\vx_t, t) \|_2^2 + 2 \Delta_u(\vx_t, t)^\top \tilde \beta_t \epsilon + \| \tilde \beta_t \epsilon \|_2^2 - \| \tilde \beta_t \epsilon \|_2^2    \right] \\
    & = - \frac{1}{2 \tilde \beta_t^2} \| \Delta_u(\vx_t, t) \|_2^2 - \frac{1}{\tilde \beta_t} \Delta_u(\vx_t, t)^\top \epsilon.
\end{align}
For the full RND we obtain
\begin{align}
    \label{eq:discrete_iw_ddpm_coef}
     \sum_{t=1}^T \log \left( \frac{p_\theta^\text{data}(\vx_{t-1}|\vx_t)}{p_\varphi^u(\vx_{t-1}|\vx_t)} \right) &= \sum_{t=1}^T \left[ -\frac{1}{2} \tilde \beta_t^{-2} \| \Delta_u(\vx_t, t) \|_2^2 + \tilde \beta_t^{-1} \Delta_u(\vx_t, t)^\top \epsilon \right] \\
     &= \sum_{t=1}^T \left[ -\frac{1}{2} \gamma_t \| u_\phi(\mathbf{x}t, t) \|_2^2 - \sqrt{\gamma_t} u_\phi(\mathbf{x}_t, t)^\top \epsilon \right],
\end{align}
with $\gamma_t = \frac{\beta_t}{\alpha_t(1 - \bar \alpha_{t-1})}$, which mimics a discretised version of the continuous RND in Equation~\eqref{eq:rnd_final}.

\section{Proof to Theorem \ref{the:}}
\label{app:proof_resampling}
\begin{proof}
Without loss of generality, we assume $\lambda=1$ for the proof.

For the first inequality in \eqref{eq:the_claim}, we observe that $\sP^{u^{k+1}}$ corresponds to the path measure of the SDE \eqref{eq:forward_SDE} with initial condition $\bX_0\sim \tilde p_{u^{k}}=\tilde p_{u^k}$. Hence, Lemma~\ref{lemma:sol_sde} implies that for any trajectory $\vx_{[0:T]}$ it holds
$$
\frac{\dd \sP^{u^{k+1}}}{\dd \sP^\mathrm{data}}(\vx_{[0:T]})=\frac{p_{u^{k+1}}(\vx_0)}{p_\mathrm{data}(\vx_0)}=\frac{\tilde p_{u^{k}}(\vx_0)}{p_\mathrm{data}(\vx_0)}
$$
Hence, it follows that
\begin{align}
\mathrm{KL}(\sP^{u^{k+1}}||\sP^\mathrm{data})&=\mathbb{E}_{\vx_{[0:T]}\sim\sP^{u^{k+1}}}\left[\ln\left(\frac{\dd \sP^{u^{k+1}}}{\dd \sP^\mathrm{data}}(\vx_{[0:T]})\right)\right]\\
&=\mathbb{E}_{\vx_0\sim\tilde p_{u^k}}\left[\ln\left(\frac{\tilde p_{u^{k}}(\vx_0)}{p_\mathrm{data}(\vx_0)}\right)\right]=\mathrm{KL}(\tilde p_{u^k}||p_\mathrm{data})\leq \mathrm{KL}(\tilde \sP^{u^k}||\sP^\mathrm{data}),
\end{align}
where the last inequality is known as data-processing inequality stating that the KL divergence of two distributions is always lower-bounded by the KL divergence of their marginals.
In particular, since the marginals of $\sP^{u^{k+1}}$ and $\tilde sP^{u^k}$ at time zero coincide, we get
\begin{align}
\mathcal F(\sP^{u^{k+1}})&=\mathrm{KL}(\sP^{u^{k+1}}||\sP^\mathrm{data})+\mathbb{E}_{\vx_{[0:T]}\sim\sP^{u^{k+1}}}[r(\vx_0)]\\
&\leq \mathrm{KL}(\tilde \sP^{u^k}||\sP^\mathrm{data})+\mathbb{E}_{\vx_{[0:T]}\sim\tilde \sP^{u^k}}[r(\vx_0)]=\mathcal F(\tilde \sP^{u^k}).
\end{align}
This shows the first inequality in \eqref{eq:the_claim}.

For the second one, we note that Bayes theorem implies $\frac{\dd \tilde{\sP}^{u^k}}{\dd  \sP^{u^k}}(\vx_{[0,T]}) = \frac{\alpha_k(\vx_{[0,T]})}{\mathbb{E}_{\vx_{[0,T]}\sim\sP^{u^k}}[\alpha_k(\vx_{[0,T]})]}$
Then, we following similar ideas as \cite[Prop 8 (ii)]{hertrich2024importance}: Let $Z=\mathbb{E}_{\vx_{[0,T]}\sim \sP^{u^k}}[\alpha(\vx_{[0,T]})]$.
We can estimate by Jensens' inequality that
\begin{align*}
-\log(Z)&=-\log\left(\mathbb{E}_{\vx_{[0,T]}\sim \sP^{u^k}}\left[\min\left(\frac{\dd \sP^\mathrm{data}}{\dd\sP^{u^k}}(\vx_{[0,T]})\frac{\exp(r(\vx_0))}{c},1\right)\right]\right)\\
&\leq -\mathbb{E}_{\vx_{[0,T]}\sim \sP^{u^k}}\left[\log\left(\min\left(\frac{\dd \sP^\mathrm{data}}{\dd\sP^{u^k}}(\vx_{[0,T]})\frac{\exp(r(\vx_0))}{c},1\right)\right)\right]\\
&=\mathbb{E}_{\vx_{[0,T]}\sim \sP^{u^k}}\left[\max\left(\log\left(\frac{\dd \sP^{u^k}}{\dd\sP^\mathrm{data}}(\vx_{[0,T]})\frac{c}{\exp(r(\vx_0))}\right),0\right)\right]
\end{align*}
On the other side, we have by definition that
\begin{align*}
\mathcal F(\tilde \sP^{u^k})&=
\mathrm{KL}(\tilde{\sP}^{u^k},\sP^\mathrm{data})-\mathbb{E}_{\vx_{[0:T]}\sim\tilde{\sP}^{u^k}}[r(\vx_0)]\\
&=\mathbb{E}_{\vx_{[0:T]}\sim\tilde{\sP}^{u^k}}\left[\log\left(\frac{cZ}{\exp(r(\vx_0))}\frac{\dd \tilde{\sP}^{u^k}}{\dd\sP^\mathrm{data}}(\vx_{[0,T]})\right)\right]-\log(cZ).
\end{align*}
By taking the expectation over $\sP^{u^k}$ instead of $\tilde{\sP}^{u^k}$ this is equal to
\begin{align}
\mathbb{E}_{\vx_{[0:T]}\sim\sP^{u^k}}\left[\frac{\dd \tilde{\sP}^{u^k}}{\dd \sP^{u^k}}(\vx_{[0,T]})\log\left(\frac{cZ}{\exp(r(\vx_0))}\frac{\dd \tilde{\sP}^{u^k}}{\dd\sP^{u^k}}(\vx_{[0,T]})\frac{\dd\sP^{u^k}}{\dd\sP^\mathrm{data}}(\vx_{[0,T]})\right)\right]-\log(cZ). \label{eq:proof_formula}
\end{align}
Inserting the formula from Bayes theorem and then the definition of $\alpha_k$, the part inside the expectation can be rewritten as
\begin{align*}
&\quad\frac{\alpha_k(\vx_{[0,T]})}{Z}\log\left(\frac{c}{\exp(r(x))}\alpha_k(\vx_{[0,T]})\frac{\dd\sP^{u^k}}{\dd\sP^\mathrm{data}}(\vx_{[0,T]})\right)\\
&=\frac{\alpha(\vx_{[0,T]})}{Z}\log\left(\frac{c}{\exp(r(x))}\min\left(1,\frac{\dd \sP^\text{data}}{\dd \sP^{u^k}}(\vx_{[0,T]}) \frac{\exp(r(\vx_0))}{c}\right)\frac{\dd\sP^{u^k}}{\dd\sP^\mathrm{data}}(\vx_{[0,T]})\right)\\
&=\frac{\alpha_k(\vx_{[0,T]})}{Z}\min\left(\log\left(\frac{\dd\sP^{u^k}}{\dd\sP^\mathrm{data}}(\vx_{[0,T]})\frac{c}{\exp(r(\vx_0))}\right), 0\right)\\
&=\frac1Z\min\left(\alpha_k(\vx_{[0,T]})\log\left(\frac{\dd\sP^{u^k}}{\dd\sP^\mathrm{data}}(\vx_{[0,T]})\frac{c}{\exp(r(\vx_0))}\right), 0\right)\\
&=\frac1Z\min\!\left(\!\min\!\left(\!1,\frac{\dd \sP^\text{data}}{\dd \sP^{u^k}}(\vx_{[0,T]}) \frac{\exp(r(\vx_0))}{c}\!\right)\!\log\!\left(\frac{\dd\sP^{u^k}}{\dd\sP^\mathrm{data}}(\vx_{[0,T]})\frac{c}{\exp(r(\vx_0))}\right), 0\!\right).
\end{align*}
Since $1\leq \frac{\dd \sP^\text{data}}{\dd \sP^{u^k}}(\vx_{[0,T]}) \frac{\exp(r(\vx_0))}{c}$ if and only if $\log\left(\frac{\dd\sP^{u^k}}{\dd\sP^\mathrm{data}}(\vx_{[0,T]})\frac{c}{\exp(r(\vx_0))}\right)\leq 0$ the minimum is attained either for both $\min$ in the above formula in the first argument or it is attained for both $\min$ in the second argument. Consequently inserting the above formula into \eqref{eq:proof_formula} is equal to
\begin{align*}
\frac1Z\mathbb{E}_{\vx_{[0:T]}\sim\sP^{u^k}}\left[\min\left(\log\left(\frac{\dd\sP^{u^k}}{\dd\sP^\mathrm{data}}(\vx_{[0,T]})\frac{c}{\exp(r(\vx_0))}\right), 0\right)\right]-\log(cZ)\\
\leq \mathbb{E}_{\vx_{[0:T]}\sim\sP^{u^k}}\left[\min\left(\log\left(\frac{\dd\sP^{u^k}}{\dd\sP^\mathrm{data}}(\vx_{[0,T]})\frac{c}{\exp(r(\vx_0))}\right), 0\right)\right]-\log(Z)-\log(c),
\end{align*}
where the inequality comes from the fact that $Z\in(0,1]$ and that the expectation is non-positive (since the integrand is non-positive). Inserting the formula of $-\log(Z)$ from the beginning of the proof, this is equal to
\begin{align*}
&\quad\mathbb{E}_{\vx_{[0:T]}\sim\sP^{u^k}}\left[\min\left(\log\left(\frac{\dd\sP^{u^k}}{\dd\sP^\mathrm{data}}(\vx_{[0,T]})\frac{c}{\exp(r(\vx_0))}\right), 0\right)\right]\\
&\qquad+\mathbb{E}_{\vx_{[0,T]}\sim \sP^{u^k}}\left[\max\left(\log\left(\frac{\dd \sP^{u^k}}{\dd\sP^\mathrm{data}}(\vx_{[0,T]})\frac{c}{\exp(r(\vx_0))}\right),0\right)\right]-\log(c)\\
&=\mathbb{E}_{\vx_{[0:T]}\sim\sP^{u^k}}\left[\log\left(\frac{\dd\sP^{u^k}}{\dd\sP^\mathrm{data}}(\vx_{[0,T]})\frac{c}{\exp(r(\vx_0))}\right)\right]-\log(c)\\
&=\mathbb{E}_{\vx_{[0:T]}\sim\sP^{u^k}}\left[\log\left(\frac{\dd\sP^{u^k}}{\dd\sP^\mathrm{data}}(\vx_{[0,T]})\right)\right]-\mathbb{E}_{\vx_{[0:T]}\sim\sP^{u^k}}[r(\vx_0)]\\
&=\mathrm{KL}(\sP^{u^k},\sP^\mathrm{data})-\mathbb{E}_{\vx_{[0:T]}\sim\sP^{u^k}}[r(\vx_0)].
\end{align*}
This shows the second inequality in \eqref{eq:the_claim} and finishes the proof.
\end{proof}

\section{Replay Buffers as Moving Average}\label{app:replay}

As outlined in Section~\ref{sec:training_network} we are using replay buffers for the numerical implementation of the importance fine-tuning algorithm. Once the buffer is

\paragraph{Replacing Random Samples from the Buffer}
In the case, that we are replace random samples from the buffer, it can be interpreted as moving average method and argue that $\mathcal F(\sP^{u^k})$ still decreases when using them.

To this end, let $u^k$ be the estimation of the control in step $k$ and recall that $\sP^{u^k}$ is the path distribution of the backward SDE \eqref{eq:controlled_SDE} with control $u^k$. Since $u^k$ was trained with the fine-tuning loss \eqref{eq:supervised_finetuning_loss}, the samples in the replay buffer are distributed as $\sP^{u^k}_0$.
Now let $\tilde\sP^{u^k}$ be the accepted samples in step $k+1$. Then, the trajectories corresponding to the samples in the replay buffer at time $k+1$ are distributed as $(1-\gamma)\sP^{u^k}_0+\gamma\tilde \sP^{u^k}_0$, where $\gamma$ is the number of accepted samples at time $k+1$ divided by the buffer size.
At the same time, we know by the convexity of $\mathcal F$ in the linear space of measures (up to a constant it coincides with the KL divergence) that
$$
\mathcal F((1-\gamma)\sP^{u^k}+\gamma\tilde \sP^{u^k})\leq (1-\gamma)\mathcal F(\sP^{u^k})+\gamma\mathcal F(\tilde \sP^{u^k})=\mathcal F(\sP^{u^k})-\gamma(\mathcal F(\sP^{u^k})-\mathcal F(\tilde \sP^{u^k}))\leq \mathcal F(\sP^{u^k}),
$$
where the last inequality follows from Theorem~\ref{the:} (ii).
In particular, following the same arguments as Theorem~\ref{the:} (iii), we obtain that the minimiser $u^{k+1}$ of the fine-tuning loss \eqref{eq:supervised_finetuning_loss} using a dataset with the distribution $(1-\gamma)\sP^{u^k}_0+\gamma\tilde \sP^{u^k}_0$
fulfils $\mathcal F(\sP^{u^{k+1}})\leq \F((1-\gamma)\sP^{u^k}+\gamma\tilde \sP^{u^k})\leq \mathcal F(\sP^{u^{k}})$.

\paragraph{Replacing the Oldest Samples from the Buffer}
In practice, we typically substitute the oldest samples from the buffer rather than choosing them at random. This is based on the intuition that more recent samples are produced by an improved model and thus are more in line with the target distribution.
Although we cannot derive a similar justification as we do for random replacement, we have observed numerically that both approaches yield comparable final results while replacing the oldest samples results in faster convergence.

\section{Experimental Details and Additional Results}\label{app:exp_details}

\subsection{Posterior Sampling for Inverse Problems}\label{app:posterior}
We train a unconditional score-based diffusion model on the \texttt{Flowers} dataset with the VP-SDE with $\beta_\text{min}=0.1$ and $\beta_\text{max}=20.0$. We parametrise the score model using a small Attention UNet \cite{dhariwal2021diffusion} with approx.~$10$M parameters. We parametrise the control as in Equation~\eqref{eq:lkhd_informed_model} with approx. $0.8$M parameters. We used $50$ importance fine-tuning iterations with $20$ gradient updates per iteration. For every iteration we sampled $64$ new trajectories and used a resampling rate of $50\%$. We used a maximal buffer size of $256$ images and no KL-regularisation. The total training time is approximately $8$min on a single NVIDIA GeForce RTX~4090.

\begin{figure}[t]
    \centering
    \includegraphics[width=1.0\linewidth,trim=100 120 55 100, clip]{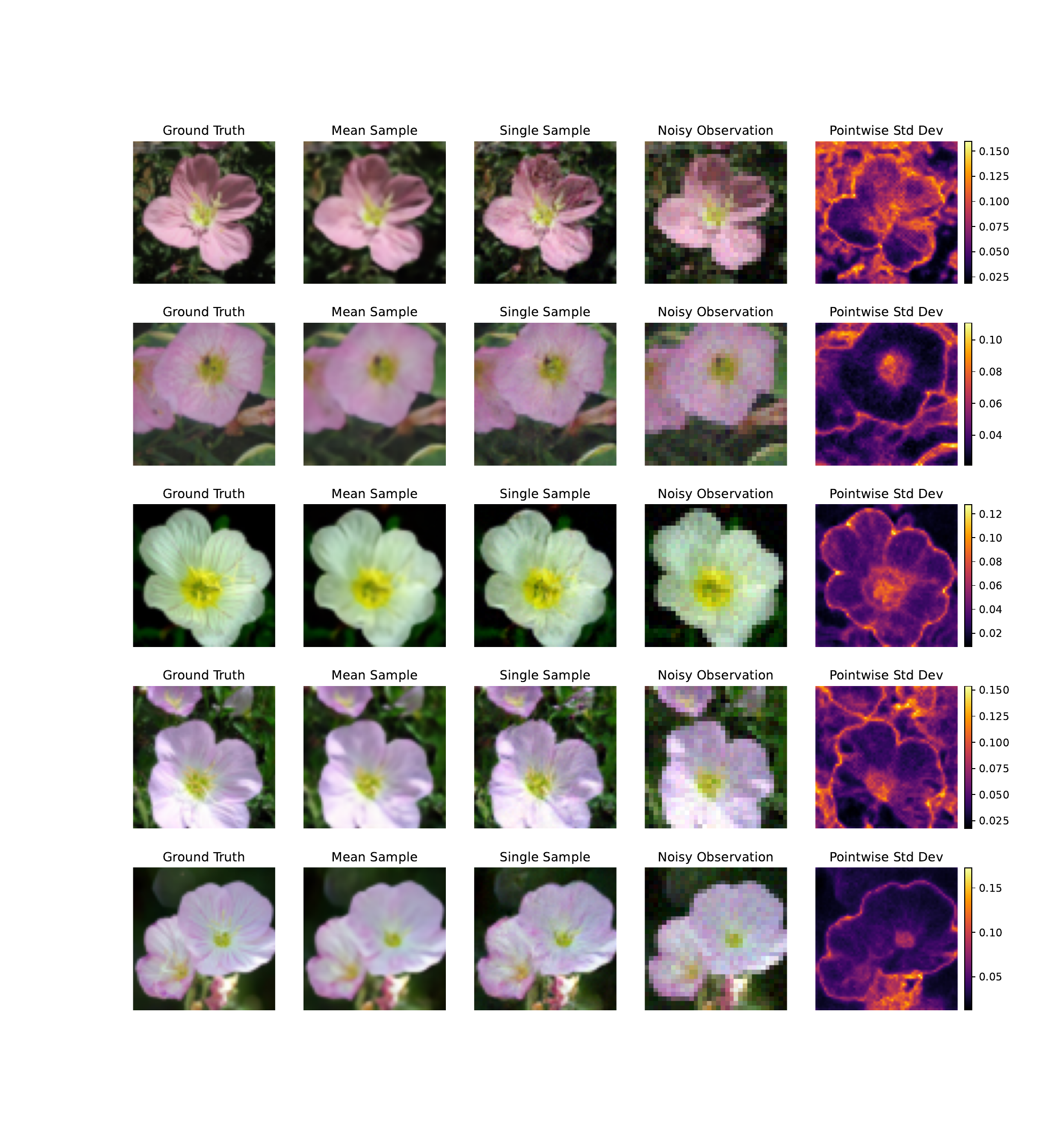}
    \caption{Additional results for different images from the test set of the \texttt{Flowers} dataset.}
    \label{fig:flower_additional_results}
\end{figure}

For the first five test images, we draw $64$ posterior samples per image and compute the per-pixel mean and standard deviation. The resulting visualisations are shown in Figure~\ref{fig:flower_additional_results}. The standard deviation is highest near object boundaries, where uncertainty is expected, and remains low in homogeneous background regions. Across all images, the uncertainty remains stable and does not collapse, indicating that fine-tuning preserves sample diversity while enforcing measurement consistency.

\subsection{Reward Fine-tuning}
\label{app:rewardfinetuning}
For alignment, we utilise \texttt{ImageReward-v1.0} \cite{xu2024imagereward}, a model trained to predict human preference rewards $r(\vx; c)$ for a given image $\vx$ and prompt $c$. We learn the tilted distribution for a specific prompt and directly fine-tune the model utilising Low-Rank Adaptation (LoRA) \cite{hu2021lora}.

Our training configuration is as follows:
\begin{itemize}
    \item \textbf{Model Architecture:} LoRA applied to all attention layers with a rank of $10$.
    \item \textbf{Sampling Buffer:} We maintain a buffer of $64$ images and sample $128$ new images per iteration. The acceptance threshold $c$ is chosen to accept the top $10\%$ of images.
    \item \textbf{Optimisation:} We train for $25$ outer iterations. Each outer iteration consists of $40$ gradient descent steps with an effective batch size of $64$. 
    \item \textbf{Loss and Regularisation:} We use a KL-regularisation weight of $2.5$ (Eqn. \eqref{eq:kl_reg}). We employ the AdamW optimiser \cite{loshchilov2017decoupled} with a weight decay of $0.001$, a starting learning rate of $1 \times 10^{-4}$, and cosine decay to $1 \times 10^{-5}$.
    \item \textbf{Evaluation:} Final evaluation uses the DDIM scheduler with $50$ time steps and a guidance scale of $5.0$.
\end{itemize}
Training was performed on a single NVIDIA GeForce RTX 4090 and completed in approximately $7$ hours.

\textbf{Top-K Sampling} is a widely known approach to increase the performance of text-to-image diffusion models without additional training. Here, we sample a batch of $K$ images and only return the image with the highest reward. In our implementation we use $K=6$. Top-K sampling is easy to implement and requires no additional training loss, but often decrease diversity and increased the sampling time by a factor of $K$. 

\textbf{FK-Steering} was proposed by \cite{singhal2025a} as a general inference-time approach to control diffusion models. We use the implementation by the authors using thee hyperparameters from the original publication\footnote{\url{https://github.com/zacharyhorvitz/Fk-Diffusion-Steering}}. The original implementation of FK-Steering used $k=4$ particles for Stable Diffusion. However, we used $k=6$ to be consistent with the Top-K sampling baseline. We observe slightly higher scores for $k=6$, which is consistent with the observations in \cite{singhal2025a}. 

\textbf{DPOK} is an approach to fine-tune diffusion models using a differentiable reward \cite{fan2024reinforcement}. We use the implementation by the authors using the provided hyperparameters\footnote{\url{https://github.com/google-research/google-research/tree/master/dpok}}. DPOK uses LoRA for the fine-tuning parametrisation. Fine-tuning takes $28$h on a single A100.

\textbf{Adjoint Matching} reformulates fine-tuning as a stochastic optimal control problem and casting it as a regression problem. We use the implementation by the authors with the provided hyperparameters\footnote{\url{https://github.com/microsoft/soc-fine-tuning-sd}}. Adjoint Matching fine-tunes the full model and does not use LoRA. This requires large GPUs and the experiments cannot be performed on a NVIDIA Geforce RTX 4090. We use the adjoint matching variant which uses CFG to sample trajectories, but not in the adjoint computation. This setting is not covered by theory, but leads to the best empirical results.   

\textbf{Relative Trajectory Balance (RTB)} proposed by \cite{venkatraman2024amortizing} arises from a GFlowNet perspective on diffusion models. We use the implementation by the authors with the provided hyperparameters\footnote{\url{https://github.com/GFNOrg/diffusion-finetuning}}.

\paragraph{Computational Resources}
Table \ref{tab:runtime_memory_stablediff} reports approximate training times and peak GPU memory usage for the different methods. Note that DPOK and Adjoint Matching were executed on a machine with an A100 GPU. These numbers should be interpreted cautiously, as training time and GPU memory can be traded off, for example, via gradient accumulation to increase effective batch size or gradient checkpointing. The main purpose of this comparison is to illustrate that our approach can be run on smaller GPUs, whereas the default configurations of DPOK or Adjoint Matching generally require larger GPUs.

\begin{table}[t]
\caption{Comparison of run-time and peak GPU memory for reward fine-tuning.}
\resizebox{\textwidth}{!}{%
\begin{tabular}{lccccc}
 \toprule                    & Base Model & DPOK & RTB & Adjoint Matching & Importance FT \\ \midrule
Training Time $(h)$    & N/A        & 28   & 17  & 4                & 7             \\
Peak GPU Memory (GB) & 9          & 34   & 20  & 36               & 16           \\ \bottomrule
\end{tabular}}
\label{tab:runtime_memory_stablediff}
\end{table}

\begin{figure}
    \centering
    \includegraphics[width=1.0\linewidth]{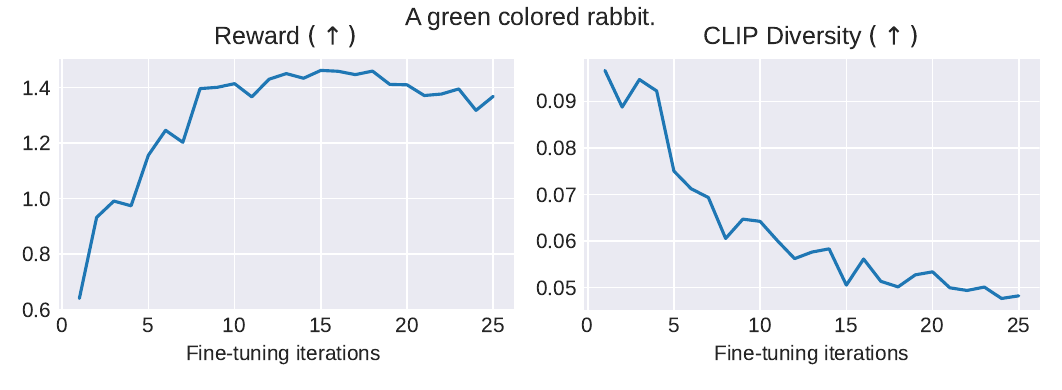}
    \caption{Reward and diversity for text-to-image fine-tuning on "A green colored rabbit." during training. As known for LoRA fine-tuning the diversity decreases over iterations.}
    \label{fig:rewarddiversity_train_rabbit}
\end{figure}

\begin{figure}[t]
\begin{subfigure}[t]{.19\textwidth}
    \includegraphics[width=0.5\linewidth]{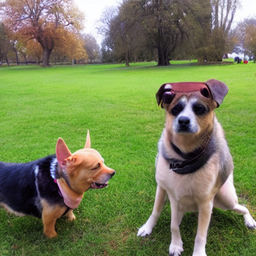}%
    \includegraphics[width=0.5\linewidth]{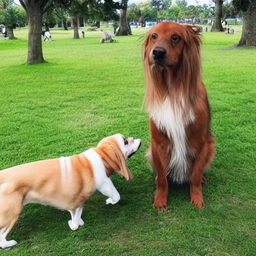}
\end{subfigure}%
\hfill
\begin{subfigure}[t]{.19\textwidth}
    \includegraphics[width=0.5\linewidth]{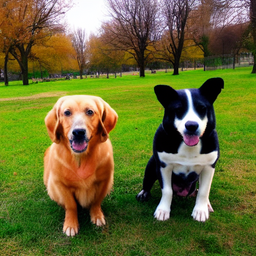}%
    \includegraphics[width=0.5\linewidth]{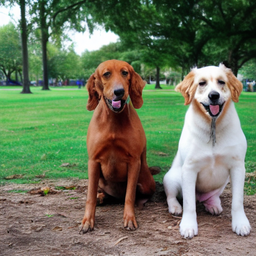}
\end{subfigure}%
\hfill
\begin{subfigure}[t]{.19\textwidth}
    \includegraphics[width=0.5\linewidth]{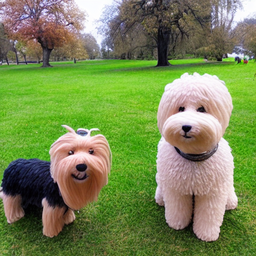}%
    \includegraphics[width=0.5\linewidth]{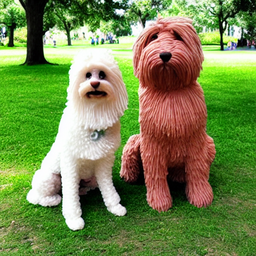}
\end{subfigure}%
\hfill
\begin{subfigure}[t]{.19\textwidth}
    \includegraphics[width=0.5\linewidth]{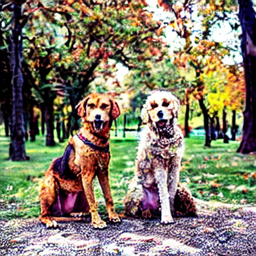}%
        \includegraphics[width=0.5\linewidth]{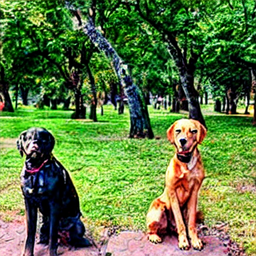}
\end{subfigure}%
\hfill
\begin{subfigure}[t]{.19\textwidth}
    \includegraphics[width=0.5\linewidth]{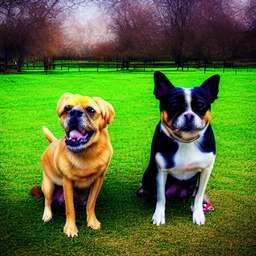}%
        \includegraphics[width=0.5\linewidth]{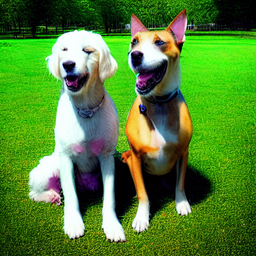}
\end{subfigure}%

\begin{subfigure}[t]{.19\textwidth}
    \includegraphics[width=0.5\linewidth]{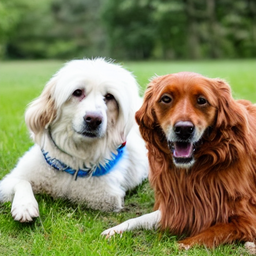}%
    \includegraphics[width=0.5\linewidth]{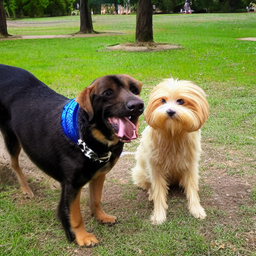}
     \captionsetup{justification=centering}
    \caption*{Base Model}
\end{subfigure}%
\hfill
\begin{subfigure}[t]{.19\textwidth}
    \includegraphics[width=0.5\linewidth]{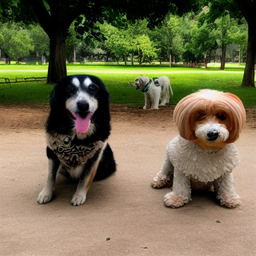}%
    \includegraphics[width=0.5\linewidth]{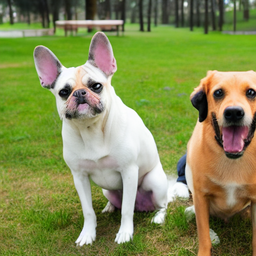}
    \captionsetup{justification=centering}
  \caption*{DPOK}
\end{subfigure}%
\hfill
\begin{subfigure}[t]{.19\textwidth}
    \includegraphics[width=0.5\linewidth]{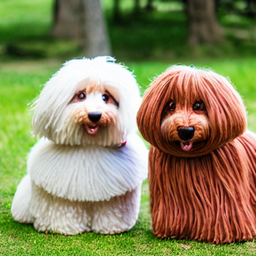}%
    \includegraphics[width=0.5\linewidth]{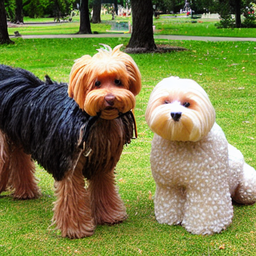}
    \captionsetup{justification=centering}
  \caption*{RTB}
\end{subfigure}%
\hfill
\begin{subfigure}[t]{.19\textwidth}
    \includegraphics[width=0.5\linewidth]{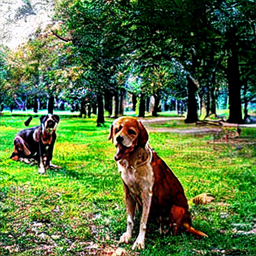}%
        \includegraphics[width=0.5\linewidth]{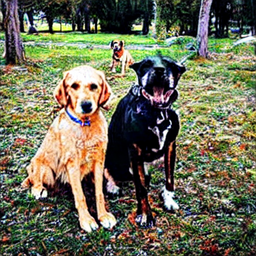}
    \captionsetup{justification=centering}
  \caption*{Adjoint Matching}
\end{subfigure}%
\hfill
\begin{subfigure}[t]{.19\textwidth}
    \includegraphics[width=0.5\linewidth]{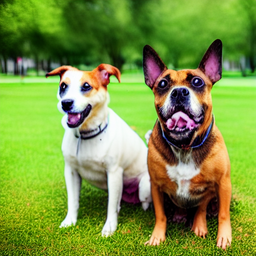}%
        \includegraphics[width=0.5\linewidth]{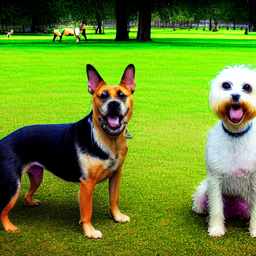}
    \captionsetup{justification=centering}
  \caption*{Importance FT}
\end{subfigure}%
\caption{Samples for the base model, DPOK, Adjoint Matching and our importance FT for the prompt ''Two dogs in the park.``. Images were generated using the same seed.} \label{fig:stable_diff_dogs}
\end{figure}

\paragraph{Additional Results}
We provide supplementary qualitative results in the following figures:
\begin{itemize}
    \item Figures~\ref{fig:stable_diff_roses} and \ref{fig:stable_diff_dogs} present samples for additional prompts.
    \item Figure~\ref{fig:rewarddiversity_train_rabbit} plots the evolution of reward and diversity metrics over the course of training iterations for the prompt ``A green colored rabbit''.
\end{itemize}

\begin{figure}[t]
\begin{subfigure}[t]{.19\textwidth}
    \includegraphics[width=0.5\linewidth]{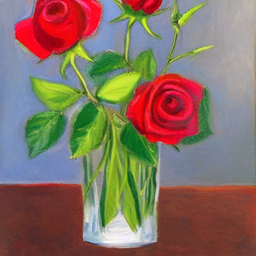}%
    \includegraphics[width=0.5\linewidth]{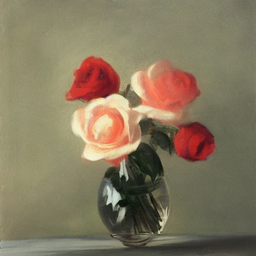}
\end{subfigure}%
\hfill
\begin{subfigure}[t]{.19\textwidth}
    \includegraphics[width=0.5\linewidth]{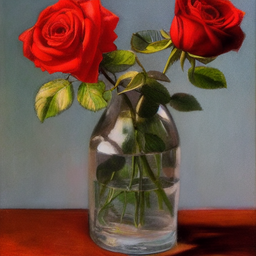}%
    \includegraphics[width=0.5\linewidth]{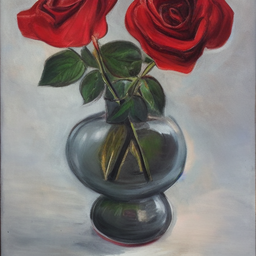}
\end{subfigure}%
\hfill
\begin{subfigure}[t]{.19\textwidth}
    \includegraphics[width=0.5\linewidth]{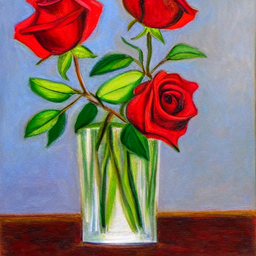}%
    \includegraphics[width=0.5\linewidth]{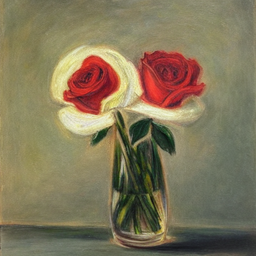}
\end{subfigure}%
\hfill
\begin{subfigure}[t]{.19\textwidth}
    \includegraphics[width=0.5\linewidth]{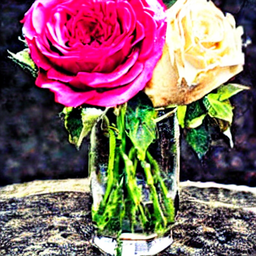}%
        \includegraphics[width=0.5\linewidth]{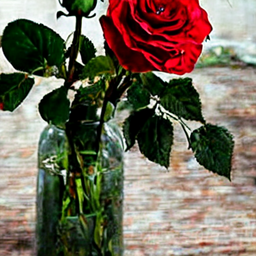}
\end{subfigure}%
\hfill
\begin{subfigure}[t]{.19\textwidth}
    \includegraphics[width=0.5\linewidth]{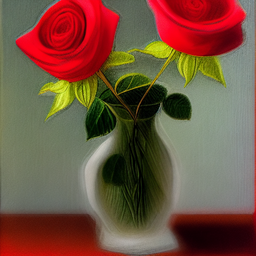}%
        \includegraphics[width=0.5\linewidth]{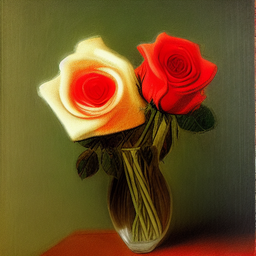}
\end{subfigure}%

\begin{subfigure}[t]{.19\textwidth}
    \includegraphics[width=0.5\linewidth]{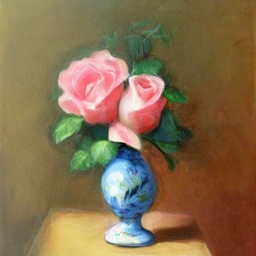}%
    \includegraphics[width=0.5\linewidth]{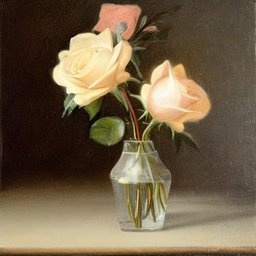}
     \captionsetup{justification=centering}
    \caption*{Base Model}
\end{subfigure}%
\hfill
\begin{subfigure}[t]{.19\textwidth}
    \includegraphics[width=0.5\linewidth]{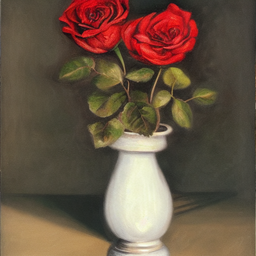}%
    \includegraphics[width=0.5\linewidth]{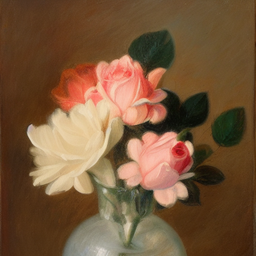}
    \captionsetup{justification=centering}
  \caption*{DPOK}
\end{subfigure}%
\hfill
\begin{subfigure}[t]{.19\textwidth}
    \includegraphics[width=0.5\linewidth]{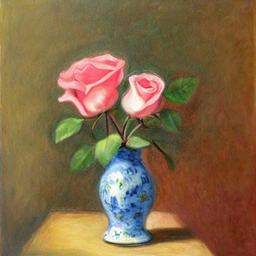}%
    \includegraphics[width=0.5\linewidth]{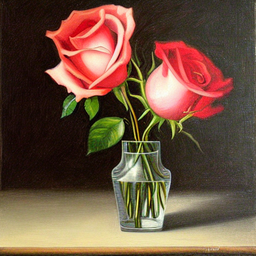}
    \captionsetup{justification=centering}
  \caption*{RTB}
\end{subfigure}%
\hfill
\begin{subfigure}[t]{.19\textwidth}
    \includegraphics[width=0.5\linewidth]{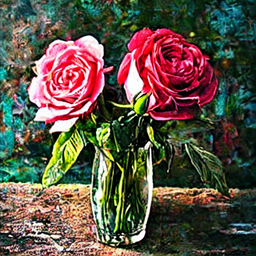}%
        \includegraphics[width=0.5\linewidth]{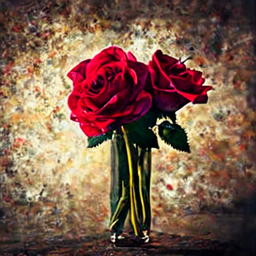}
    \captionsetup{justification=centering}
  \caption*{Adjoint Matching}
\end{subfigure}%
\hfill
\begin{subfigure}[t]{.19\textwidth}
    \includegraphics[width=0.5\linewidth]{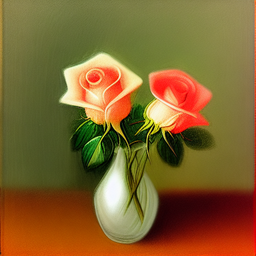}%
        \includegraphics[width=0.5\linewidth]{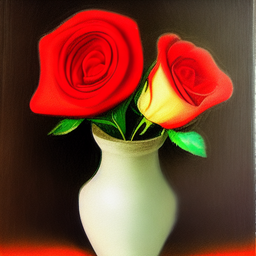}
    \captionsetup{justification=centering}
  \caption*{Importance FT}
\end{subfigure}%
\caption{Samples for the base model, DPOK, Adjoint Matching and our importance FT for the prompt ''Two roses in a vase.``. Images were generated using the same seed.} \label{fig:stable_diff_roses}
\end{figure}

\begin{figure}[t]
\begin{subfigure}[t]{.48\textwidth}
\begin{subfigure}[t]{.48\textwidth}
    \includegraphics[width=0.5\linewidth]{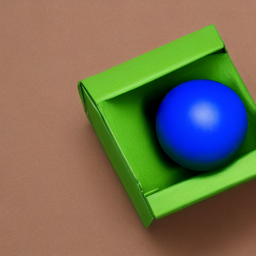}%
    \includegraphics[width=0.5\linewidth]{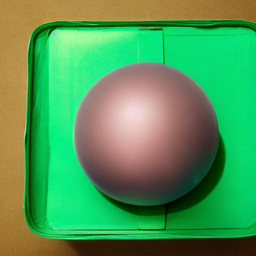}
\end{subfigure}%
\hfill
\begin{subfigure}[t]{.48\textwidth}
    \includegraphics[width=0.5\linewidth]{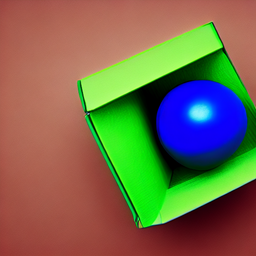}%
        \includegraphics[width=0.5\linewidth]{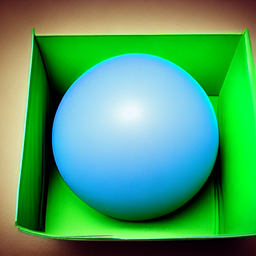}
\end{subfigure}%

\begin{subfigure}[t]{.48\textwidth}
    \includegraphics[width=0.5\linewidth]{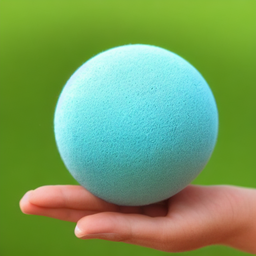}%
    \includegraphics[width=0.5\linewidth]{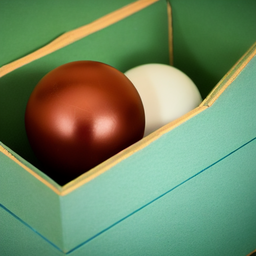}
     \captionsetup{justification=centering}
    \caption*{Base Model\\(mean reward $-0.18$)}
\end{subfigure}%
\hfill
\begin{subfigure}[t]{.48\textwidth}
    \includegraphics[width=0.5\linewidth]{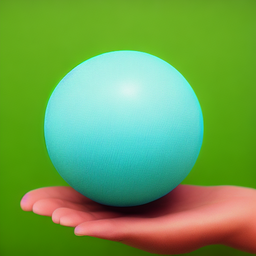}%
        \includegraphics[width=0.5\linewidth]{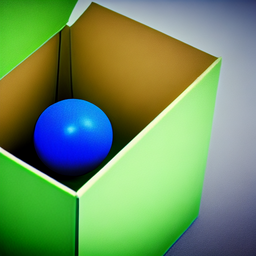}
    \captionsetup{justification=centering}
  \caption*{Importance FT\\(mean reward $1.02$)}
\end{subfigure}%
\caption*{''A blue ball in a green box``}
\end{subfigure}
\hfill
\begin{subfigure}[t]{.48\textwidth}
\begin{subfigure}[t]{.48\textwidth}
    \includegraphics[width=0.5\linewidth]{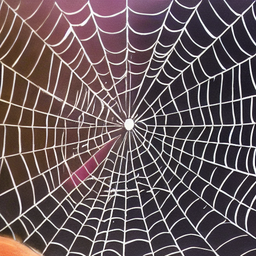}%
    \includegraphics[width=0.5\linewidth]{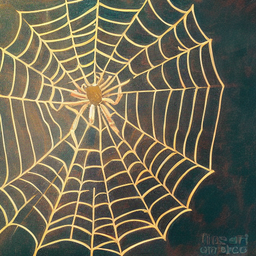}
\end{subfigure}%
\hfill
\begin{subfigure}[t]{.48\textwidth}
    \includegraphics[width=0.5\linewidth]{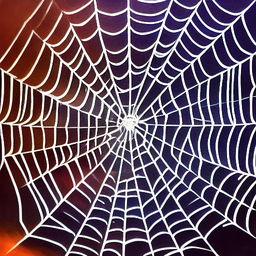}%
        \includegraphics[width=0.5\linewidth]{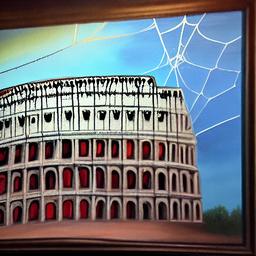}
\end{subfigure}%

\begin{subfigure}[t]{.48\textwidth}
    \includegraphics[width=0.5\linewidth]{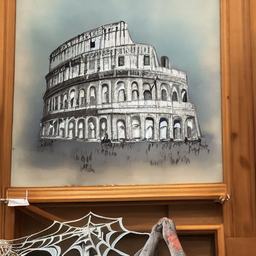}%
    \includegraphics[width=0.5\linewidth]{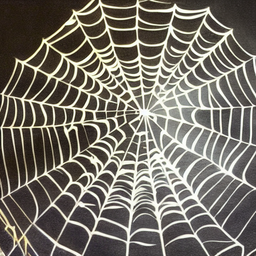}
     \captionsetup{justification=centering}
    \caption*{Base Model\\(mean reward $-1.15$)}
\end{subfigure}%
\hfill
\begin{subfigure}[t]{.48\textwidth}
    \includegraphics[width=0.5\linewidth]{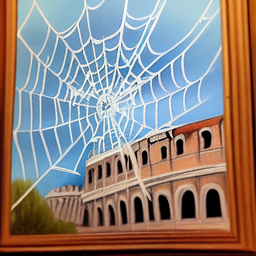}%
        \includegraphics[width=0.5\linewidth]{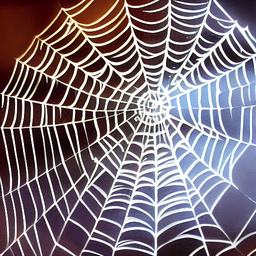}
    \captionsetup{justification=centering}
  \caption*{Importance FT\\(mean reward $0.18$)}
\end{subfigure}%
\caption*{''A spider web and a painting of the Colosseum``}
\end{subfigure}
\caption{Additional prompts for importance fine-tuning.} \label{fig:additional}
\end{figure}

\clearpage

\bibliographystyle{siamplain}
\bibliography{references}

\end{document}